\documentclass[11pt,a4paper]{article}

\usepackage[T1]{fontenc}
\usepackage[utf8]{inputenc}

\usepackage[a4paper, left=2cm, right=2cm, top=2.5cm, bottom=2.5cm]{geometry}

\usepackage{amsmath}
\usepackage{amssymb}
\usepackage{amsthm}
\usepackage{bm}

\usepackage{graphicx}
\usepackage{xcolor}
\usepackage{subcaption}
\usepackage{rotating}

\usepackage{booktabs}
\usepackage{multirow}
\usepackage{tabularx}
\usepackage{adjustbox}
\usepackage{threeparttable}

\usepackage{algorithm}
\usepackage{algpseudocode}

\usepackage{placeins}
\usepackage{float}

\usepackage{pifont}
\usepackage{makecell}

\newcommand{\BHHO}{\text{BHHO}}
\newcommand{\BGWO}{\textsc{BGWO}}
\newcommand{\ABHG}{\textsc{ABHG}}
\newcommand{\DE}{\textsc{DE}}

\usepackage{tikz}
\usetikzlibrary{
    shapes.geometric,
    arrows.meta,
    positioning,
    3d,
    backgrounds
}

\usepackage[numbers,sort&compress]{natbib}

\usepackage[
    colorlinks=true,
    linkcolor=blue,
    citecolor=blue,
    urlcolor=blue,
    breaklinks=true
]{hyperref}

\title{
EMG-Hybrid: Elite Memory Gravity-Guided Dual-Engine Metaheuristic for High-Dimensional Feature Selection
}
\author{
    Saba Zare$^{1}$,\ Mohammad Taheri$^{1,*}$
    \\[1.5ex]
    {\small $^{1}$Department of Electrical and Computer Engineering,
    University of Shiraz, Shiraz, Iran}
    \\[1ex]
    {\small $^{*}$Corresponding author: \texttt{motaheri@shirazu.ac.ir}}\\
    {\small ORCID: \href{https://orcid.org/0000-0003-4888-5883}{0000-0003-4888-5883}}
    \\[1ex]
    {\small Email (first author): \texttt{sabazare.zrz@gmail.com}}
}
\date{}

\begin{document}

\maketitle


\begin{abstract}
Wrapper-based feature selection in noisy, high-dimensional datasets is challenging because the binary search space grows exponentially, increasing the risk of diversity loss and premature convergence. Existing Harris Hawks--Grey Wolf hybrids use fixed schedules, limiting adaptation to changing search conditions. This paper presents EMG-Hybrid, a three-level hierarchical metaheuristic integrating search control, candidate generation, and elite preservation. Level~1 uses a dual-engine, state-dependent controller. The macro $\Lambda$-Engine uses normalized per-locus Shannon entropy to govern transitions between binary Harris Hawks-inspired exploration (BHHO) and binary Grey Wolf-inspired exploitation (BGWO). The micro $\mathcal{M}$-Engine adjusts a bounded, phase-aware acceptance threshold based on the convergence velocity of the best archived fitness. At Level~2, an Adaptive Black Hole Gravity (ABHG) kernel performs elite-guided binary assimilation from rank-selected archive members, at a rate decreasing with Hamming distance and fitness gap. At Level~3, a write-protected top-$k$ archive preserves the best-so-far solutions for refinement through a discrete differential evolution kernel with strictly greedy selection. Refined solutions are periodically reinserted at an evaluation cost of at most 2.5\%. Experiments on 14 UCI and OpenML datasets employed a $k$-NN wrapper, leakage-free nested cross-validation, and 2,000 fitness evaluations. Compared with eight metaheuristics and Forward Selection, EMG-Hybrid achieved the best average ranks in accuracy (1.61) and fitness (2.36). Accuracy improvements over all eight metaheuristics were significant after Holm correction. Subset sizes were competitive, but not always smallest. Ablations highlighted the contributions of diversity-guided switching, multi-solution archiving, adaptive acceptance, and ABHG guidance.
\end{abstract}


\maketitle
\section{Introduction}\label{sec1}

Feature selection can be formulated as a search over a $d$-dimensional binary space, where each dimension denotes the inclusion (1) or exclusion (0) of a feature. With $2^d$ possible subsets, the search complexity grows exponentially in $d$, causing a severe combinatorial explosion that hinders the discovery of informative feature subsets. This challenge is compounded by noisy or interdependent features and is especially severe in high-dimensional data analytics, such as image recognition, text classification, and genomics, where the number of features often far exceeds the number of samples. Without effective dimension reduction, predictive models tend to overfit, incur greater computational cost, and exhibit reduced accuracy due to irrelevant features that obscure the true signal and redundant features that increase the mathematical burden~\cite{Chandrashekar2014Survey}.

Filter methods, wrapper methods, and embedded methods are the three main categories into which feature selection approaches can generally be divided. Regardless of any learning algorithm, filter approaches rank or choose features based on statistical or information-theoretic criteria. They are computationally efficient but often overlook intricate interdependencies among features. Embedded methods incorporate feature selection directly into the model training process, leading to faster solutions but making the selected features tightly coupled with the biases of the specific learning algorithm and its regularization. The role and necessity of individual features may also remain obscure in such approaches due to implicit interactions during training~\cite{Chandrashekar2014Survey}.

Wrapper methods, in contrast, evaluate candidate feature subsets by training and validating a learning algorithm on each subset. Better prediction performance is usually the outcome of this intimate integration, particularly when classification accuracy is the primary goal~\cite{Liu2012Feature}. However, since the evaluation depends directly on the classifier's output, the selected features tend to be closely linked to the inductive biases of the specific learning model as well. Thus, highly efficient search strategies are crucial to handle the exponentially growing subset space within advanced AI methods and applications.

Various search strategies have been employed within wrapper frameworks. Greedy algorithms follow straightforward iterative procedures but can easily become trapped in local optima, particularly in high-dimensional settings. Random search improves scalability by sampling the feature space stochastically, yet lacks a systematic mechanism to refine solutions progressively~\cite{Thaher2020BinaryHHO}.

To overcome these challenges, researchers increasingly rely on metaheuristic algorithms as the core search engines within wrapper-based feature selection frameworks. Inspired by natural processes such as evolutionary adaptation, swarm intelligence, and thermal annealing, these methods effectively balance global exploration of the feature space with local exploitation of promising regions~\cite{Blum2003Metaheuristics}. This capability allows them to effectively navigate non-convex, fragmented, and high-dimensional landscapes where gradient-based techniques often fail. Well-known instances of metaheuristic algorithms are Genetic Algorithms (GA)~\cite{Goldberg1989GA}, Ant Colony Optimization (ACO)~\cite{Dorigo2004ACO}, and Particle Swarm Optimization (PSO)~\cite{Kennedy1995PSO}, which have been extensively utilized to select minimal yet informative feature sets, ultimately improving the performance of predictive models.

In recent years, the Grey Wolf Optimizer (GWO) and Harris Hawks Optimization (HHO) have gained traction as competitive alternatives for wrapper-based feature selection. GWO emulates the leadership dynamics and coordinated hunting tactics of grey wolves, guiding the search through elite solutions ($\alpha$, $\beta$, and $\delta$)~\cite{Mirjalili2014GWO}. In contrast, HHO models the cooperative ambush strategies of Harris’s hawks, adjusting its behavior across four distinct modes depending on the ``energy'' of escaping prey~\cite{Heidari2019HHO}. Both algorithms are adapted for binary feature selection by applying binarization mechanisms that convert continuous updates into binary decisions. 
Despite their advantages, both GWO and HHO are known to suffer from premature convergence and a rapid loss of population diversity when applied to high-dimensional binary search spaces, and their wrapper-based evaluation is computationally expensive. Hybrid designs such as the framework of Al-Wajih et al.~\cite{HBGWOHHO_2021} combine HHO for global exploration with GWO for local exploitation, but such hybrids typically rely on predefined, iteration-based transitions between the two search mechanisms. A fixed schedule cannot react to the state of the population, which may lead to premature exploitation or to wasted evaluations on an already homogeneous population.

Recent surveys of metaheuristic feature selection identify the balance between exploration and exploitation, the preservation of promising solutions, and the design of effective binary search operators as persistent challenges in high-dimensional optimization~\cite{dokeroglu2022survey}. In binary feature selection, purely stochastic updates may disrupt promising feature combinations, while excessive reliance on current elite solutions reduces diversity and encourages premature convergence. These challenges motivate state-dependent search control, elite-conditioned candidate generation, and a protected memory of high-quality subsets.

To address these issues, this study introduces EMG-Hybrid (Elite Memory Gravity Hybrid), a three-level hierarchical framework for wrapper-based feature selection that separates search control, candidate generation, and elite preservation:

\begin{itemize}
\item \textbf{Level 1 --- Dual-Engine Control Layer.}
The macro $\Lambda$-Engine computes the normalized per-locus Shannon entropy of the population as a diversity index. When this index falls below a threshold $\tau$ and a minimum phase inertia has elapsed, it switches the search between binary Harris Hawks-inspired exploration (BHHO) and binary Grey Wolf-inspired exploitation (BGWO). The micro $\mathcal{M}$-Engine tracks the convergence velocity of the best archived fitness and adapts a bounded acceptance threshold $\theta_t$: the threshold is enlarged under stagnation and in the exploration phase, and reduced in the exploitation phase. The two engines are coupled in one direction only, since $\theta_t$ never influences the phase.

\item \textbf{Level 2 --- Unified Search Operator.}
Each generation applies a single mode-conditioned operator. The active phase (BHHO or BGWO) produces a continuous step on a bipolar encoding of the binary population, which is mapped back to binary through a sigmoid transfer function and a Bernoulli draw. An Adaptive Black Hole Gravity (ABHG) kernel then copies differing bits from a rank-biased archived elite into a random subset of individuals, with a rate that decreases with their Hamming distance to the elite and with their fitness gap. A $1/d$ bit-flip mutation, an empty-subset repair, and an elite clone follow. A candidate replaces its parent if its fitness is not worse than the parent's fitness plus $\theta_t$. All archive queries at this level are read-only.

\item \textbf{Level 3 --- Elite Memory Archive.}
The archive stores the $k$ best unique subsets ranked by the scalar fitness (error and subset size combined by a weighted sum), so no Pareto dominance is used. Levels~1 and~2 cannot modify it; it is updated only by Level~3 from the evaluated population in every generation, so the best-so-far solution is never lost. At a fixed interval of main-search evaluations, a discrete Differential Evolution (DE) kernel refines the archived elites under strictly greedy selection, with a mutation mask driven by the Level~1 threshold, and the refined elites are reinserted into the population. 
\end{itemize}

The parameters of EMG-Hybrid are listed in Table~\ref{tab:parameter_settings_ranges}, and the sensitivity of the main ones is analyzed in Section~\ref{sec:sensitivity}. The population size, the evaluation budget, and the fitness weights are inputs of the method.

EMG-Hybrid was evaluated on 14 UCI and OpenML benchmark datasets using leakage-free nested cross-validation and a budget of 2,000 fitness evaluations. It was compared with eight metaheuristic baselines and Forward Selection where results were available. EMG-Hybrid achieved the best mean ranks in classification accuracy (1.61) and fitness (2.36), with significantly higher accuracy than all eight baselines after Holm correction. For subset size, EHHO (2.14) and BGWOHHO (2.57) achieved better mean ranks than EMG-Hybrid (3.14), although their differences were not statistically significant. Ablation experiments further examined the contributions of the framework's main components.

The subsequent parts of this paper are arranged as follows. Section \ref{sec2} reviews related metaheuristic approaches for feature selection. The proposed hybrid framework and its adaptive mechanisms are described in Section~\ref{sec4}. The experimental design, findings, and analysis are presented in Section~\ref{sec5}. Future research directions are outlined and concluded in Section~\ref{sec6}.


\section{Related Work}\label{sec2}

This section reviews metaheuristic wrapper-based feature selection from the perspective of the challenges that arise in high-dimensional, binary search spaces, rather than as a catalogue of algorithm families. Section~\ref{sec2.1} summarizes the main families, Section~\ref{sec2.2} discusses hybrid, adaptive, and large-scale strategies, Section~\ref{sec2.3} analyzes the open challenges, and Section~\ref{sec2.4} derives the research gap addressed by EMG-Hybrid.

\subsection{Metaheuristic Feature Selection}\label{sec2.1}

Metaheuristics are widely used as search engines in wrapper-based feature selection because they explore large, non-convex subset spaces and balance global exploration with local exploitation. The main families are summarized below, together with representative recent variants.

\textbf{PSO-based methods.} The Altruistic Adaptive PSO (AAPSO) combines altruistic actions with adaptive inertia-weight adjustment and was applied to biomedical imaging~\cite{Pramanik2022AAPSO}. A hybrid PSO for text feature selection refines the velocity update and uses chaotic-map initialization to increase population diversity and reduce premature convergence~\cite{Nachaoui2024PSO}.

\textbf{ACO-based methods.} A two-stage hybrid ACO (TSHFS-ACO) first estimates the optimal subset size and then applies a relevance-guided wrapper~\cite{TSHFS_ACO_2021}. A clustering-based ACO combines Laplacian scores with K-means silhouette indices to improve feature relevance and diversity~\cite{Clustering_ACO_2023}.

\textbf{GA-based methods.} Genetic algorithms remain popular owing to their natural binary encoding. Recent work hybridizes GAs with filter-based population initialization, using rank-based and random sampling filters to improve accuracy and shorten convergence time on microarray data~\cite{Abasabadi2022GaRankRand}.

\textbf{GWO-based methods.} Variants address convergence and local-optima entrapment through a self-repulsion strategy (GWO-SRS)~\cite{GWO_SRS_2025}, nonlinear parameter control with neighborhood mutation (AMGWO)~\cite{AMGWO_2025}, and a dispersion-based random walk (RWGWO) for chronic-disease prediction~\cite{RWGWO_2022}.

\textbf{HHO-based methods.} A hierarchical HHO (EHHO) with binary encoding and multi-level search was proposed to improve accuracy and runtime~\cite{EHHO_2023}.

\subsection{Hybrid, Adaptive, and Large-Scale Strategies}\label{sec2.2}

\textbf{Algorithm hybrids.} BGWOHHO uses HHO for exploration and GWO for exploitation in binary feature selection~\cite{HBGWOHHO_2021}. A related architecture lets GWO lead the search while HHO refines locally, but it targets global optimization benchmarks rather than feature selection~\cite{Tu2023}. A GWO framework combining elite learning and chaotic initialization with a genetic algorithm was applied to hyperspectral feature selection~\cite{Shang2025GWOGA}. Black-hole-based search has been adapted to binary feature selection~\cite{Pashaei2017BBHA}, and the Modified Binary BHO (MBHO) adds inversion mutation and a correlation-based fitness to improve diversity and accuracy~\cite{AlEiadeh2024}.

\textbf{Elite archives.} Elite mechanisms appear in the Elite Hybrid Binary Cuckoo Search (EHBCS) for high-dimensional classification~\cite{Zhao2021EHBCS} and in an elite-GA-enhanced LSTM that reports lower RMSE and compact subsets in earthquake prediction~\cite{ega2024lstm}.

\textbf{Memetic and local-refinement schemes.} Combining global search with local refinement is the defining idea of memetic algorithms~\cite{moscato1989evolution,neri2012memetic}. In a standard memetic scheme, a local search operator is applied directly to individuals of the evolving population. Vignolo and Gerard~\cite{Vignolo2026MOELIGA} propose MOELIGA, a multi-objective genetic algorithm in which the best chromosomes seed subordinate populations that are evolved within the active feature subspace of each parent and then compared with it. MOELIGA also uses a crowding-based fitness-sharing mechanism, a sigmoid-shaped cardinality objective, and a geometry-based class-separability objective, and it was evaluated on 14 datasets with up to roughly 20{,}000 features. The authors report that the local-improvement stage increases the computational cost and that the advantage diminishes on small-sample, multi-class data such as Yeoh. In this scheme, the main population is paused while the subordinate populations evolve, so refinement is interleaved with the main search rather than isolated from it.

\textbf{Feature grouping and multi-stage search-space reduction.} A recent line of work reduces the effective search space before or during the metaheuristic search. Ye et al.~\cite{Ye2026FGIHBO} first filter features with mRMR, then partition the retained features into low-redundancy subspaces using symmetric uncertainty, and optimize them with an improved Hybrid Breeding Optimization algorithm (FGIHBO). FGIHBO adds GWO-guided updates and a simulated-annealing acceptance rule whose cooling rate is adapted from the Shannon entropy of the population. They evaluate it on eleven gene-expression datasets with 2{,}000--12{,}533 features and a $k$-NN wrapper. In this design, entropy regulates the annealing temperature inside a single optimizer; it is not used to time transitions between separate search regimes. TSHFS-ACO~\cite{TSHFS_ACO_2021} and filter-initialized GAs~\cite{Abasabadi2022GaRankRand} follow the same multi-stage logic, with a filter or subset-size estimation stage ahead of the wrapper.

Table~\ref{tab:related} compares the methods most relevant to this study in terms of their representations, evaluation strategies, elite-preservation mechanisms, and applicability to high-dimensional feature selection. The final column identifies the comparison dimension relevant to EMG-Hybrid; it is not intended to imply limitations or shortcomings reported by the original authors.

\begin{table}[htbp]
\caption{Comparison of representative feature-selection methods
and the proposed EMG-Hybrid.}
\label{tab:related}
\centering
\small
\renewcommand{\arraystretch}{1.2}
\setlength{\tabcolsep}{5pt}

\begin{tabularx}{\textwidth}{
@{}
>{\raggedright\arraybackslash}p{2.4cm}
>{\raggedright\arraybackslash}X
>{\raggedright\arraybackslash}p{2.1cm}
>{\raggedright\arraybackslash}X
@{}
}
\toprule
Method & Main mechanism & Evaluation & Elite strategy \\
\midrule

AAPSO~\cite{Pramanik2022AAPSO}
& PSO-based adaptive inertia and altruistic learning
& Binary, $k$-NN
& None \\

GWO-SRS~\cite{GWO_SRS_2025}
& GWO-based self-repulsion and elite head-wolf learning
& Binary, classifier-based
& Elite head-wolf \\

AMGWO~\cite{AMGWO_2025}
& GWO-based nonlinear control and adaptive neighborhood mutation
& Binary, classifier-based
& None \\

EHHO~\cite{EHHO_2023}
& HHO-based hierarchical population organization
& Binary, $k$-NN
& None \\

BGWOHHO~\cite{HBGWOHHO_2021}
& Hybrid HHO exploration and GWO exploitation
& Binary, $k$-NN
& None \\

MBHO~\cite{AlEiadeh2024}
& BHO-based inversion mutation and correlation-aware fitness
& Binary, classifier-based
& None \\

EHBCS~\cite{Zhao2021EHBCS}
& Cuckoo search with Relief weighting and GA operators
& Binary, SVM
& Elite preservation \\

FGIHBO~\cite{Ye2026FGIHBO}
& HBO-GWO hybrid with mRMR/SU grouping and entropy-adaptive annealing
& Binary, $k$-NN
& None \\

MOELIGA~\cite{Vignolo2026MOELIGA}
& GA-based local improvement, fitness sharing, and multi-objective refinement
& Binary, classifier-independent
& Parent replacement \\

\midrule
\textbf{EMG-Hybrid}
& \textbf{BHHO-BGWO switching, adaptive acceptance, ABHG bit assimilation, and archive-level DE refinement}
& \textbf{Binary, $k$-NN}
& \textbf{Write-protected top-$k$ archive} \\

\bottomrule
\end{tabularx}
\end{table}

\subsection{Challenges in High-Dimensional Wrapper-Based Feature Selection}\label{sec2.3}

\textbf{Premature convergence and loss of diversity.} Binary search spaces grow as $2^d$, and population diversity tends to degrade quickly as $d$ increases. Existing work counters this mainly through diversity-oriented operators: chaotic initialization~\cite{Nachaoui2024PSO}, self-repulsion~\cite{GWO_SRS_2025}, neighborhood mutation~\cite{AMGWO_2025}, inversion mutation~\cite{AlEiadeh2024}, and crowding-based fitness sharing~\cite{Vignolo2026MOELIGA}. Entropy-based signals have recently been used to regulate search intensity~\cite{Ye2026FGIHBO}, but, as far as the reviewed work shows, within a single optimizer rather than as a trigger for changing the search regime.

\textbf{Exploration--exploitation imbalance.} Hybrids such as BGWOHHO assign exploration and exploitation to different algorithms~\cite{HBGWOHHO_2021,Tu2023}. As discussed in the Introduction, the transition between the two is typically governed by predefined or iteration-based scheduling, which is not conditioned on the observed state of the search.

\textbf{Preservation and refinement of high-quality sub-schemas.} Elite mechanisms have been used in cuckoo search~\cite{Zhao2021EHBCS} and GA-based pipelines~\cite{ega2024lstm}, and elite learning has been combined with GWO~\cite{Shang2025GWOGA}. MOELIGA refines elite subsets through subordinate populations~\cite{Vignolo2026MOELIGA}, but the refinement pauses the main search and compares each refined subset with its parent. Within the reviewed work, refining a protected archive of elites separately from the active population, with periodic reinsertion and an explicit evaluation cost, is less explored than direct population-level local search.

\textbf{Cost of wrapper evaluation and search-space reduction.} Each candidate subset requires training and validating a classifier, so cost grows quickly with $d$. Multi-stage and grouping approaches shrink the effective search space~\cite{TSHFS_ACO_2021,Abasabadi2022GaRankRand,Clustering_ACO_2023,Ye2026FGIHBO}. They typically rely on a filter or grouping stage whose quality bounds what the wrapper can recover; for example, discarding informative features during pre-filtering is reported to degrade accuracy~\cite{Ye2026FGIHBO}. Approaches that keep the full binary space but concentrate the search on promising regions during the run, under a controlled evaluation budget, remain comparatively less studied.

\textbf{Compact yet informative subsets.} Several methods report smaller subsets alongside accuracy gains~\cite{GWO_SRS_2025,ega2024lstm}. MOELIGA addresses the trade-off explicitly through a multi-objective formulation~\cite{Vignolo2026MOELIGA}, whereas most single-objective wrappers control it only through the fitness weights.

\subsection{Research Gap and Motivation}
\label{sec2.4}

The reviewed literature suggests four research gaps that remain insufficiently addressed in combination:
\begin{enumerate}
\item \textbf{State-dependent search allocation.} Exploration and exploitation are commonly coordinated through predefined schedules or heuristic rules. A mechanism that monitors population diversity directly and uses it to time transitions between separate search regimes, rather than only to tune a parameter of a single optimizer, remains a relevant direction.
\item \textbf{Controlled acceptance and selection pressure.} Acceptance rules are usually either strictly greedy or governed by a fixed or schedule-based tolerance. Adapting the tolerance to the observed convergence velocity of the best solution, and tying it to the search phase, has received less attention.
\item \textbf{Archive-level refinement.} Local refinement is commonly applied to the evolving population. Refining a write-protected archive of elites with strictly greedy selection, and returning the result through periodic reinsertion at a bounded evaluation cost, is less explored in high-dimensional binary spaces.
\item \textbf{Coordination of search and memory.} Hybrid methods often combine multiple search operators, but the roles of search control, candidate generation, and long-term preservation of high-quality solutions are not always explicitly separated, which makes the effect of each component hard to attribute.
\end{enumerate}

EMG-Hybrid addresses these gaps through a three-level coordination architecture. Its contribution is not the combination of BHHO and BGWO alone, nor the memetic hybridization principle itself, but the mechanisms governing their interaction: (i) a dual-engine controller in which the macro $\Lambda$-Engine uses the normalized per-locus Shannon entropy of the population to trigger switches between BHHO and BGWO, while the micro $\mathcal{M}$-Engine adapts a bounded acceptance threshold from the convergence velocity of the best archived fitness; (ii) the Adaptive Black Hole Gravity (ABHG) kernel, which assimilates bits from a rank-biased elite at a rate that decreases with the Hamming distance and the fitness gap between candidate and elite; (iii) a write-protected, fitness-ranked top-$k$ archive that is refined periodically by a discrete differential evolution kernel and reinserted into the population, at an evaluation overhead of at most $k/m_{\mathrm{FE}}=2.5\%$; and (iv) a leakage-free nested cross-validation evaluation on benchmark datasets, including high-dimensional feature-selection tasks, complemented by comparisons with baseline methods, component ablations, and a comparison of archive-level and population-coupled DE.


\section{Proposed EMG-Hybrid}\label{sec4}
This section introduces the EMG-Hybrid framework from a unified architectural perspective. To improve conceptual clarity and reduce algorithmic fragmentation, the method is organized into a three-level hierarchy.

\paragraph{Conventions.} The index $t=1,2,\dots$ denotes the \emph{generation} (one pass of Levels 1--3). $\mathbf{X}_t$ is the population at the start of generation $t$ and $\mathbf{X}_t^{+}$ the evaluated population at its end. $\mathcal{E}_t$ is the archive at the start of generation $t$. All budgets are counted in \emph{fitness evaluations} (FE): $FE_{\max}$ is the total budget, $FE_{\mathrm{main}}$ counts population evaluations (including the initial one), and the DE refinement evaluations are counted separately. The search progress is $\omega_t = FE/FE_{\max}\in[0,1]$, measured at the start of generation $t$. Every schedule in the method (BGWO decay, DE period) is expressed through FE or $\omega_t$, so no separate iteration budget $T$ is needed.

\begin{itemize}

    \item \textbf{Level 1 (Control Layer: $\Lambda + \mathcal{M}$ Engines):}
    The control layer governs the execution dynamics through a dual-signal feedback mechanism. The macro $\Lambda$-engine performs entropy-based detection of population homogenization to trigger transitions between global exploration and local exploitation, while the micro $\mathcal{M}$-engine regulates the selection pressure from the fitness trajectory of the archive.

    The objective~\cite{yang1998feature} balances classification accuracy and subset compactness:
    \begin{equation}
        \min_{\mathbf{x} \in \{0,1\}^d\setminus\{\mathbf{0}\}} f(\mathbf{x}) = \eta \cdot E(\mathbf{x}) + \mu \cdot \frac{1}{d}\sum_{i=1}^{d} x_i ,
        \label{eq:fitness}
    \end{equation}
    where $E(\mathbf{x})\in[0,1]$ is the error rate of the wrapper classifier and $\eta,\mu\in[0,1]$ are weights with $\eta+\mu=1$ ($\eta=0.99$, $\mu=0.01$; Table~\ref{tab:parameter_settings_ranges}). Hence $f(\mathbf{x})\in[0,1]$. All-zero vectors are never evaluated: they are repaired by Eq.~\eqref{eq:repair} (Level 2) or rejected (Level 3).

    \item \textbf{Level 2 (Unified Search Engine):}
    The population is evolved by a single mode-conditioned operator that integrates \BHHO{} (exploration), \BGWO{} (elite-guided exploitation), and the Adaptive Black Hole Gravity (\ABHG{}) kernel, which injects elite bits probabilistically.

    \item \textbf{Level 3 (Elite Memory System):}
    A write-protected archive of the $k$ best unique solutions supports top-down knowledge transfer. Level 3 also hosts a discrete Differential Evolution (\DE{}) sub-kernel that periodically refines the archived elites under strictly greedy selection, so refinement never degrades the archive.

\end{itemize}

\
\subsection{Level 1: Dual-Engine Control System (Decision Layer)}
\label{subsec:hybrid_switching}

Instead of epoch-based scheduling, the framework uses a dual-engine feedback controller that monitors two complementary signals: (i) structural population diversity, measured by the normalized per-locus Shannon entropy in the $\Lambda$-Engine (Eq.~\eqref{eq:diversity_index}), and (ii) functional convergence velocity, derived from the best-fitness trajectory of the archive in the $\mathcal{M}$-Engine (Eq.~\eqref{eq:stagnation_delta}). Level 1 therefore reads the current population bits (for $DI_t$) and the best archived fitness (for $\Delta_t$); it never writes to either.

The $\Lambda$-Engine detects population homogenization and triggers discrete phase transitions $\phi_t\in\{\BHHO,\BGWO\}$. The $\mathcal{M}$-Engine adjusts an acceptance threshold $\theta_t$ from stagnation conditions. The coupling is one-directional: $\theta_t$ never influences $\phi_t$, whereas the $\mathcal{M}$-Engine update reads the current phase (Eq.~\eqref{eq:theta_update}). Both signals are broadcast to Level 2 as $\mathcal{C}_t=\{\phi_t,\theta_t\}$.

\subsubsection{Entropy-Driven Phase Transition Protocol ($\Lambda$-Engine)}
\label{subsub:entropy_engine}

Let the population at generation $t$ consist of $N$ binary
vectors $\mathbf{x}^{(i)}_t\in\{0,1\}^d$. The selection
frequency of feature $j$ is defined as
\begin{equation}
    p_{t,j} =
    \frac{1}{N}\sum_{i=1}^{N}x^{(i)}_{t,j},
    \qquad j=1,\ldots,d.
    \label{eq:bit_freq}
\end{equation}
The raw population diversity is measured by the mean
per-feature binary Shannon entropy:
\begin{equation}
    H_t =
    -\frac{1}{d}\sum_{j=1}^{d}
    \left[
    p_{t,j}\log_2 p_{t,j}
    +(1-p_{t,j})\log_2(1-p_{t,j})
    \right],
    \label{eq:raw_diversity}
\end{equation}
where $0\log_2 0=0$. However, the maximum attainable value
of this measure depends on the population's feature-selection
density. Consequently, raw entropy values may not be directly
comparable across datasets with substantially different
selection densities.

To address this limitation, the $\Lambda$-Engine normalizes
the raw diversity by the entropy associated with the
population's mean selection density. Let
\begin{equation}
    \bar{p}_t =
    \frac{1}{d}\sum_{j=1}^{d}p_{t,j},
    \qquad
    H(\bar{p}_t) =
    -\bar{p}_t\log_2\bar{p}_t
    -(1-\bar{p}_t)\log_2(1-\bar{p}_t).
    \label{eq:mean_selection_entropy}
\end{equation}
The normalized Diversity Index is then defined as
\begin{equation}
    DI_t =
    \frac{H_t}{H(\bar{p}_t)}.
    \label{eq:diversity_index}
\end{equation}
By the concavity of binary entropy, $DI_t\in[0,1]$
whenever $H(\bar{p}_t)>0$. A value close to one indicates
that the population retains diversity close to the maximum
attainable at its current selection density, whereas a value
close to zero indicates strong agreement among individuals
in their feature-selection patterns. This normalization
reduces the influence of differences in selection density
and provides a common diversity scale across datasets.
When $H(\bar{p}_t)=0$, the population has a uniform
selection pattern across all individuals; the index is
defined as zero.

The $\Lambda$-Engine uses the normalized index to monitor
population diversity and determine phase transitions.
Entropic stagnation is detected when
\begin{equation}
    DI_t \leq \tau,
    \label{eq:stagnation_cond}
\end{equation}
where $\tau\in(0,1)$ is a fixed threshold. In this study,
$\tau=0.5$ is used uniformly across all datasets, without
dataset-specific threshold adjustment. Thus, the switching
criterion is based on the proportion of diversity retained
relative to the maximum attainable at the current selection
density, rather than on the absolute entropy value.

Let $\phi_t\in\{\BHHO,\BGWO\}$ denote the active phase at
generation $t$, with $\phi_0=\BHHO$. The phase transition
rule is given by
\begin{equation}
    \phi_t =
    \begin{cases}
        \overline{\phi_{t-1}} &
        \text{if } DI_t\leq\tau
        \;\wedge\;
        t-t_{\mathrm{last}}\geq t_{\min},\\
        \phi_{t-1} & \text{otherwise},
    \end{cases}
    \label{eq:switching_rule}
\end{equation}
where $\overline{\phi}$ denotes the complementary phase,
$t_{\mathrm{last}}$ records the generation of the most
recent phase transition, and $t_{\min}=10$ imposes a
minimum phase duration to prevent rapid switching.
Initially, $t_{\mathrm{last}}=0$, and it is updated to $t$
after each transition.

\subsubsection{Dynamic Fitness Threshold Intensity Adjustment ($\mathcal{M}$-Engine)}
\label{subsub:fitness_engine}

The $\mathcal{M}$-Engine regulates a scalar \emph{acceptance threshold} $\theta_t\in[\theta_{\min},\theta_{\max}]$ and has no influence on the phase-switching logic.

\paragraph{Operational meaning of $\theta_t$.}
The $\mathcal{M}$-Engine regulates a dimensionless relative
acceptance coefficient $\theta_t\in[\theta_{\min},\theta_{\max}]$.
Rather than specifying an absolute fitness deterioration,
$\theta_t$ scales the permissible deterioration according
to the fitness level of the best archived solution.
Specifically, a candidate $\mathbf{x}_i'$ replaces its
parent $\mathbf{x}_i$ if
\begin{equation}
    f(\mathbf{x}_i')
    \leq
    f(\mathbf{x}_i)
    +\theta_t\max(f_{\mathrm{elite},t},10^{-3}),
    \label{eq:theta_acceptance_l2}
\end{equation}
where $f_{\mathrm{elite},t}$ denotes the best fitness
stored in the archive at the time of acceptance, and
$f\in[0,1]$ is minimized. The relative scaling makes
the acceptance tolerance responsive to the fitness
scale of each dataset. A larger $\theta_t$ permits
greater deterioration relative to the elite fitness,
thereby weakening parent--offspring selection pressure,
whereas $\theta_t\rightarrow0$ approaches strict
greedy replacement. The archive independently preserves
the best-so-far solution, even when a worse candidate
is accepted into the population.

\begin{enumerate}
    \item \textbf{Level 2 replacement rule.}
    The relative acceptance criterion in
    Eq.~\eqref{eq:theta_acceptance_l2} controls
    parent--offspring replacement.

    \item \textbf{Level 3 mutation intensity.}
    Through the map $g$ (Eq.~\eqref{eq:g_function}),
    $\theta_t$ determines the mutation mask $\psi_t$
    of the DE kernel (Eq.~\eqref{eq:mutation_update}).
    The DE selection step (Eq.~\eqref{eq:greedy_selection_de})
    remains strictly greedy.
\end{enumerate}

Using the same adaptive coefficient for both operations
provides a shared stagnation-driven control signal;
therefore, the effects of acceptance tolerance and
mutation intensity are coupled.

\paragraph{Convergence velocity.}
Let $f^*_t=f(\mathbf{x}^*_1)$ be the best archived fitness at the start of
generation $t$. Since the archive is monotone
(Eq.~\eqref{eq:elite_memory_update}), the velocity
\begin{equation}
    \Delta_t = f^*_{t-1}-f^*_{t} \ \ge 0, \qquad t\ge2,
    \label{eq:stagnation_delta}
\end{equation}
needs no absolute value. A raw stagnation counter is updated by
\begin{equation}
    \hat{s}_t =
    \begin{cases}
        s_{t-1} + 1, & \text{if } \Delta_t \le \epsilon,\\
        0, & \text{otherwise,}
    \end{cases}
    \qquad s_1 = 0,
    \label{eq:stagn_counter}
\end{equation}
(no history at $t=1$), and stagnation is declared when
$\hat{s}_t \ge S_{\mathrm{tol}}$, with $\epsilon=10^{-4}$ and
$S_{\mathrm{tol}}=5$. Note that an improvement obtained only by removing a
single feature is worth $\mu/d$ in $f$; for $d>1000$ this is below $\epsilon$
and is deliberately treated as stagnation, while a single additional correctly
classified validation sample exceeds $\epsilon$ whenever the validation set has
fewer than $10^{4}$ samples.

\paragraph{Bounded threshold update.}
After $\phi_t$ and $\hat{s}_t$ are known, the threshold used in generation $t$
is
\begin{equation}
    \theta_{t} =
    \begin{cases}
        \min\!\left(\theta_{t-1} \lambda_{\mathrm{stagn}},\ \theta_{\max}\right), & \text{if } \hat{s}_t \ge S_{\mathrm{tol}} \quad (\text{stagnation}),\\[2pt]
        \min\!\left(\theta_{t-1} \lambda_{\mathrm{up}},\ \theta_{\max}\right), & \text{if } \hat{s}_t < S_{\mathrm{tol}} \text{ and } \phi_t=\BHHO,\\[2pt]
        \max\!\left(\theta_{t-1} \lambda_{\mathrm{down}},\ \theta_{\min}\right), & \text{if } \hat{s}_t < S_{\mathrm{tol}} \text{ and } \phi_t=\BGWO.
    \end{cases}
    \label{eq:theta_update}
\end{equation}
Once $\theta_t$ has been computed, the counter is reset after a
stagnation-triggered expansion,
\begin{equation}
    s_t =
    \begin{cases}
        0, & \text{if } \hat{s}_t \ge S_{\mathrm{tol}},\\
        \hat{s}_t, & \text{otherwise.}
    \end{cases}
    \label{eq:stagn_reset}
\end{equation}
Thus $\theta_t$ always uses the counter value before the reset, and the reset
limits consecutive stagnation expansions to at most one per
$S_{\mathrm{tol}}=5$ generations.

Stagnation triggers an expansion by $\lambda_{\mathrm{stagn}}=1.05$, while
$\BHHO$ and $\BGWO$ apply gradual expansion and contraction using
$\lambda_{\mathrm{up}}=1.02$ and $\lambda_{\mathrm{down}}=0.98$, respectively.
The threshold is bounded by $\theta_{\min}=0.001$ and $\theta_{\max}=0.3$,
with initial value $\theta_0=0.01$. The threshold is retained across phase
switches, ensuring continuous adaptation without reinitialization. With a
minimum phase duration of $t_{\min}=10$ generations, the gradual updates alone
(i.e., without stagnation events) change $\theta_t$ by factors of approximately
$1.02^{10}\approx 1.22$ under $\BHHO$ and $0.98^{10}\approx 0.82$ under
$\BGWO$ over one minimum-length phase. Stagnation expansions add at most a
factor of $1.05$ per $S_{\mathrm{tol}}$ generations, and $\theta_t$ always
remains within $[\theta_{\min},\theta_{\max}]$. These bounded, incremental
adjustments allow the threshold to adapt to search conditions without abrupt
changes or unbounded growth.

\paragraph{Mutation-intensity map $g$.} The DE mutation mask is a bounded linear function of $\theta_t$:
\begin{equation}
    \psi_t = g(\theta_t) = \psi_{\min} + (\psi_{\max}-\psi_{\min})\,\frac{\theta_t-\theta_{\min}}{\theta_{\max}-\theta_{\min}},
    \label{eq:g_function}
\end{equation}
with $\psi_{\min}=0.05$ and $\psi_{\max}=0.5$. Hence $g$ is increasing and $\psi_t\in[\psi_{\min},\psi_{\max}]\subset(0,1)$. DE is invoked only periodically, so each call samples the \emph{current} value $\psi_t=g(\theta_t)$ at the moment of the call: stagnation, which raises $\theta_t$, widens the perturbation span, and a long \BGWO{} phase narrows it.

\subsection{Level 2: Unified EMG Search Operator (Execution Layer)}
\label{subsec:unified_emg_operator}

To avoid a collection of heterogeneous components, all search behaviors are consolidated into one operator $\mathcal{O}_{\mathrm{EMG}}$, a mode-conditioned transformation of the population applied as an ordered composition of stages:
\begin{equation}
    \mathbf{X}_t^{+} = \mathcal{O}_{\mathrm{EMG}}(\mathbf{X}_t \mid \mathcal{C}_t, \mathcal{E}_t)
    = \mathcal{G}_{\theta_t} \circ \mathcal{K} \circ \mathcal{R} \circ \mathcal{F}_{p_{\mathrm{mut}}} \circ \mathcal{A}_{\ABHG}(\cdot \mid \mathcal{E}_t) \circ \mathcal{T} \circ \mathcal{U}_{\phi_t}(\mathbf{X}_t \mid \mathcal{E}_t),
    \label{eq:unified_operator_core}
\end{equation}
where $\mathcal{C}_t=\{\phi_t,\theta_t\}$ is the control channel from Level 1. The stages are applied right to left:
\begin{enumerate}
    \item $\mathcal{U}_{\phi_t}$: continuous phase update (\BHHO{} or \BGWO{});
    \item $\mathcal{T}$: stochastic transfer (binarization);
    \item $\mathcal{A}_{\ABHG}$: \ABHG{} bit assimilation;
    \item $\mathcal{F}_{p_{\mathrm{mut}}}$: low-rate bit-flip mutation;
    \item $\mathcal{R}$: empty-subset repair;
    \item $\mathcal{K}$: elite clone, $\mathbf{x}_N\leftarrow\mathbf{x}^*_1$;
    \item $\mathcal{G}_{\theta_t}$: threshold acceptance after evaluation (Eq.~\eqref{eq:theta_acceptance_l2}).
\end{enumerate}
Only $\mathcal{U}_{\phi_t}$ acts in continuous space; every later stage acts on binary vectors. The clone $\mathcal{K}$ is applied \emph{after} mutation and repair, so the protected copy of $\mathbf{x}^*_1$ is never altered; its fitness is known and is not re-evaluated, and it is exempt from $\mathcal{G}_{\theta_t}$ (it replaces the individual in slot $N$ unconditionally). Consequently one generation consumes $N-1$ evaluations.

\subsubsection{Continuous--Binary Interface}
\label{subsub:cont_bin_interface}

\BHHO{} and \BGWO{} are defined in continuous space, whereas the population and the archive are binary. The interface is a \emph{bipolar encoding} followed by a stochastic transfer back to $\{0,1\}^d$. For $\mathbf{x}\in\{0,1\}^d$ define
\begin{equation}
    \hat{\mathbf{y}}(\mathbf{x}) = 2\mathbf{x}-\mathbf{1}\in\{-1,+1\}^d ,
    \label{eq:bipolar}
\end{equation}
so that $\hat y_j=+1$ means ``feature $j$ selected''. Write $\hat{\mathbf{y}}_i=\hat{\mathbf{y}}(\mathbf{x}_i)$ for the current individuals and $\hat{\mathbf{y}}^*_r=\hat{\mathbf{y}}(\mathbf{x}^*_r)$ for archive members. The interface is \emph{stateless}: no continuous position is carried across generations. Each generation encodes the current binary population, produces a continuous step vector $\mathbf{v}_i\in\mathbb{R}^d$, and maps it back to binary, so the previous bit influences the next one only through $\hat y_{i,j}$ inside the update equations.

\subsubsection{Continuous Position Update $\mathcal{U}_{\phi_t}$}
\label{subsub:cont_update}

\paragraph{\BHHO{} mode (global exploration).} \BHHO{} uses \emph{only the exploration phase} of the Harris Hawks update; the escaping-energy mechanism and the besiege (exploitation) branches are not used, because local intensification is the role of the \BGWO{} phase. The name therefore denotes an HHO-inspired exploration operator, not the full HHO algorithm. For each individual $i$, draw $q\sim\mathcal{U}(0,1)$, scalars $r_1,r_2,r_3\sim\mathcal{U}(0,1)$ and a vector $\mathbf{r}_4\sim\mathcal{U}(0,1)^d$, and select a random individual $j\neq i$:
\begin{equation}
    \mathbf{v}_i =
    \begin{cases}
        \hat{\mathbf{y}}_j - r_1\,\bigl|\hat{\mathbf{y}}_j - 2r_2\,\hat{\mathbf{y}}_i\bigr|, & q \ge 0.5,\\[2pt]
        \bigl(\hat{\mathbf{y}}^*_1 - \bar{\mathbf{y}}\bigr) - r_3\,\bigl(-\mathbf{1} + 2\,\mathbf{r}_4\bigr), & q < 0.5,
    \end{cases}
    \label{eq:bhho_update}
\end{equation}
where $|\cdot|$ is element-wise, $\bar{\mathbf{y}}=\tfrac1N\sum_{i=1}^{N}\hat{\mathbf{y}}_i$ is the population mean, and the bounds of the encoding are $[-1,+1]$. The first branch disperses individuals around a random peer; the second applies a weak attraction toward the best archived solution $\mathbf{x}^*_1$. Component ranges are $v_{i,j}\in[-4,1]$ for the first branch and $[-3,3]$ for the second; the first branch is therefore biased toward $v<0$ (feature discarding), which acts as a mild sparsity prior during exploration.

\paragraph{\BGWO{} mode (elite-guided exploitation).} The three leaders are the three best archive members, $\alpha=\mathbf{x}^*_1$, $\beta=\mathbf{x}^*_2$, $\delta=\mathbf{x}^*_3$ (hence $k\ge3$). For each leader $L\in\{\alpha,\beta,\delta\}$ and each individual $i$, with $\mathbf{r}_1,\mathbf{r}_2\sim\mathcal{U}(0,1)^d$ drawn independently per leader,
\begin{align}
    \mathbf{A}^{L} &= 2a\,\mathbf{r}_1 - a\mathbf{1}, \qquad \mathbf{C}^{L}=2\mathbf{r}_2, \qquad a = 2\,(1-\omega_t),\nonumber\\
    \mathbf{y}^{L}_i &= \hat{\mathbf{y}}(L) - \mathbf{A}^{L}\odot\bigl|\mathbf{C}^{L}\odot\hat{\mathbf{y}}(L) - \hat{\mathbf{y}}_i\bigr|,
    \qquad
    \mathbf{v}_i = \tfrac{1}{3}\bigl(\mathbf{y}^{\alpha}_i+\mathbf{y}^{\beta}_i+\mathbf{y}^{\delta}_i\bigr).
    \label{eq:bgwo_update}
\end{align}
The decay coefficient $a$ follows the budget progress $\omega_t=FE/FE_{\max}$, so it is consistent with the FE-based stopping criterion. The leaders are read from $\mathcal{E}_t$ (read-only access).

\subsubsection{Transfer Function, Probability Interpretation, and Bit Update}
\label{subsub:transfer}

The step vector is first clipped, $v_{i,j}\leftarrow\mathrm{clip}(v_{i,j},-V_{\max},V_{\max})$ with $V_{\max}=4$, and mapped through an S-shaped transfer function with steepness $\kappa=1$:
\begin{equation}
    P_{i,j} = \mathcal{T}(v_{i,j}) = \frac{1}{1+\exp(-\kappa\, v_{i,j})}.
    \label{eq:transfer}
\end{equation}
$P_{i,j}\in(0,1)$ is the \emph{selection probability} of feature $j$ for individual $i$: $P_{i,j}=0.5$ corresponds to $v_{i,j}=0$ (no preference), $v_{i,j}>0$ biases toward selecting the feature, and $v_{i,j}<0$ toward discarding it. The bit update is a conditionally independent Bernoulli draw that \emph{overwrites} the previous bit:
\begin{equation}
    x_{i,j}^{(1)} =
    \begin{cases}
        1, & \text{if } \text{rand}(0,1) < P_{i,j},\\
        0, & \text{otherwise.}
    \end{cases}
    \label{eq:bernoulli_bit}
\end{equation}
Thus $\Pr[x^{(1)}_{i,j}=1\mid\mathbf{v}_i]=P_{i,j}$, and the bits of one individual are independent given $\mathbf{v}_i$. Because of the clipping, $P_{i,j}\in[\mathcal{T}(-V_{\max}),\mathcal{T}(V_{\max})]\approx[0.018,0.982]$, so no locus becomes deterministic through the transfer alone.

\subsubsection{Elite-Conditioned Probabilistic Memory Injection (\ABHG{} Kernel)}
\label{subsub:abhg}

The operator includes an Adaptive Black Hole Gravity (\ABHG{}) kernel, a \emph{complementary} stage applied after the binarization $\mathcal{T}$. It does not replace the output of $\mathcal{T}$, does not modify the archive, and performs read-only stochastic sampling of $\mathcal{E}_t$ conditioned on elite quality and structural similarity.

\paragraph{Scope of application.} Each individual $\mathbf{x}_i$ is selected independently with probability $p_{\mathrm{abhg}}$. For a selected individual, an elite $\mathbf{x}^*_r\in\mathcal{E}_t$ is drawn by rank-biased selection (Eq.~\eqref{eq:rank_prob}).

\paragraph{Assimilation rate.} Let $d_H(\cdot,\cdot)$ be the Hamming distance and $\tilde d_H = d_H(\mathbf{x}_i,\mathbf{x}^*_r)/d\in[0,1]$ its normalized form. The assimilation factor $\Gamma_i$ (renamed to avoid a clash with the transfer probability $P_{i,j}$) is
\begin{equation}
    \Gamma_i =
    \frac{1}{1 + \tilde d_H(\mathbf{x}_i, \mathbf{x}^*_r)}
    \exp\bigl(-\left|f(\mathbf{x}_i) - f(\mathbf{x}^*_r)\right|\bigr),
    \label{eq:abhg_probability}
\end{equation}
and the per-bit assimilation rate is
\begin{equation}
    \rho_i = \rho_{\max}\, \Gamma_i, \qquad \rho_{\max} \in (0, 0.2].
    \label{eq:abhg_rate}
\end{equation}
Normalizing $d_H$ by $d$ makes $\Gamma_i$ independent of the problem dimensionality; without it, $\Gamma_i$ would vanish in spaces with thousands of features. The fitness of $\mathbf{x}_i$ is its parent's evaluated fitness from the previous generation, since the candidate has not been evaluated yet at this stage.

\paragraph{Bit-level rule.} Only bits that differ from the elite are candidates:
\begin{equation}
    x_{i,j} \leftarrow
    \begin{cases}
        x^*_{r,j}, & \text{if } x_{i,j} \neq x^*_{r,j} \ \text{and}\ \text{rand}(0,1) < \rho_i, \\
        x_{i,j}, & \text{otherwise (no operation).}
    \end{cases}
    \label{eq:abhg_bit_rule}
\end{equation}
\ABHG{} contains no bit-flipping branch, so population diversity decreases gradually and in a controlled way rather than through complementation of whole chromosomes. Memory influence is regulated jointly by structural proximity and fitness similarity, which avoids deterministic attraction toward elites. With $p_{\mathrm{abhg}}=0.2$ and $\rho_{\max}=0.1$, at most about $2\%$ of the bits are expected to change per generation; the sensitivity to $p_{\mathrm{abhg}}$ and $\rho_{\max}$ is analyzed in Section~\ref{sec:sensitivity}.

\subsubsection{Binary Pipeline: Mutation, Repair, Elite Clone and Acceptance}
\label{subsub:binary_pipeline}

The binarized population is the only output of $\mathcal{U}_{\phi_t}\!\to\!\mathcal{T}$ that later stages see. The remaining stages act sequentially on binary vectors:
\begin{enumerate}
    \item \textbf{\ABHG{}} (Eqs.~\eqref{eq:abhg_probability}--\eqref{eq:abhg_bit_rule}) modifies, but does not replace, the bits drawn in Eq.~\eqref{eq:bernoulli_bit}; its output is not re-mapped through $\mathcal{T}$.
    \item \textbf{Mutation $\mathcal{F}_{p_{\mathrm{mut}}}$:} every bit of every non-clone individual is flipped independently with probability $p_{\mathrm{mut}}=1/d$.
    \item \textbf{Empty-subset repair $\mathcal{R}$:} if $\|\mathbf{x}_i\|_1=0$ after mutation, one feature is switched on, with the locus drawn by roulette selection over the transfer probabilities of that individual,
    \begin{equation}
        x_{i,j^\star}\leftarrow 1,\qquad \Pr[j^\star=j]=\frac{P_{i,j}}{\sum_{l=1}^{d}P_{i,l}} .
        \label{eq:repair}
    \end{equation}
    Roulette selection is used instead of an $\arg\max$ because \ABHG{} and mutation change bits after $P_{i,j}$ is computed, so $\arg\max_j P_{i,j}$ would deterministically favour the same locus and bias repaired vectors. $P_{i,j}$ serves only as a heuristic prior. For the initial population, where no $P_{i,j}$ exists, the locus is drawn uniformly at random.
    \item \textbf{Elite clone $\mathcal{K}$:} $\mathbf{x}_N\leftarrow\mathbf{x}^*_1$, applied after mutation and repair. The clone is non-empty by construction, keeps its known fitness, and is exempt from the acceptance rule.
    \item \textbf{Evaluation and threshold acceptance $\mathcal{G}_{\theta_t}$:} the $N-1$ non-clone candidates are evaluated, and Eq.~\eqref{eq:theta_acceptance_l2} decides whether each replaces its parent; a rejected candidate restores its parent, whose fitness is already known.
\end{enumerate}
No all-zero vector is ever evaluated or archived. The initial population is drawn with $\Pr[x_{i,j}=1]=0.5$ and repaired by Eq.~\eqref{eq:repair} with a uniform random locus.

\subsubsection{Mode-Conditioned Behavioral Dynamics}
\label{subsub:behavioral_dynamics}

The update $\mathcal{U}_{\phi_t}$ implements two behavioral regimes driven by the phase indicator $\phi_t$. When $\phi_t=\BHHO$, it induces stochastic population dispersion combined with weak elite attraction to enhance coverage of the Boolean hypercube. When $\phi_t=\BGWO$, the update is constrained by an elite-guided structure in which the search is biased toward the top-ranked solutions of $\mathcal{E}_t$ through a leadership hierarchy. Because $\mathcal{A}_{\ABHG}$ and $\mathcal{F}_{p_{\mathrm{mut}}}$ alter only a small fraction of bits (Sections~\ref{subsub:abhg}--\ref{subsub:binary_pipeline}), the structure induced by $\phi_t$ remains the dominant factor shaping the population after each generation.

\subsection{Level 3: Elite Memory Archive (Knowledge Layer)}
\label{subsec:elite_memory_archive}

The \textit{Elite Memory Archive} $\mathcal{E}_t$ is a persistent, fitness-ranked repository of the best unique solutions found so far. Unlike Level 1, which monitors the global state and regulates the search phase, and Level 2, which evolves the population, Level 3 preserves elite binary configurations and refines them incrementally.

The archive is \emph{write-protected with respect to Levels 1 and 2}: neither the decision layer nor the operator $\mathcal{O}_{\mathrm{EMG}}$ can insert, delete, or modify an archive member. Level 2 accesses $\mathcal{E}_t$ only through read-only queries (leader selection for \BGWO{}, rank-biased elite sampling for \ABHG{}, and the elite clone), and Level 1 reads only the best elite fitness $f(\mathbf{x}^*_1)$ from it. The archive is, however, \emph{not independent of the population}: it is rewritten once per generation by a dedicated Level 3 update that consumes the evaluated population $\mathbf{X}_t^{+}$ (Eq.~\eqref{eq:elite_memory_update}), and it is refined every $m_{\mathrm{FE}}$ main evaluations by the \DE{} sub-kernel. ``Write-protected'' therefore refers to the access rights of Levels 1 and 2 during their own execution phases, not to the archive's lifetime.

\subsubsection{Archive Access Protocol}
\label{subsub:archive_protocol}

Within each generation the archive is accessed in a fixed order (Table~\ref{tab:archive_access}). During Phases~1 and~2 the archive is a frozen snapshot $\mathcal{E}_t$; writes occur only in Phases~3 and~4, both executed inside Level 3. Phase~4 is performed in the same generation \emph{after} Phase~3; it does not replace a main-search generation.

\begin{table}
\centering
\caption{Archive access schedule within one generation. R = read, W = write.}
\label{tab:archive_access}
\small
\renewcommand{\arraystretch}{1.2}
\begin{tabularx}{\linewidth}{@{}c l X l@{}}
\toprule
Phase & Executed by & Operation on $\mathcal{E}$ & Mode \\
\midrule
1 & Level 1 &
Evaluate $f(\mathbf{x}^{*}_t)$ to compute $\Delta_t$
(Eq.~\eqref{eq:stagnation_delta}); compute the normalized
diversity index $DI_t$ from $\mathbf{X}_t$ &
R \\

2 & Level 2 &
Read \BGWO{} leaders and draw the \ABHG{} elite;
perform elite cloning
($\mathbf{x}_N\leftarrow\mathbf{x}^{*}_1$) &
R \\

3 & Level 3 &
Consolidate the archive,
$\mathcal{E}_t\rightarrow\mathcal{E}_{t+1}$,
using $\mathbf{X}_t^{+}$
(Eq.~\eqref{eq:elite_memory_update}) &
W (every $t$) \\

4 & Level 3 &
Refine $\mathcal{E}_{t+1}$ using \DE{} and reinsert
the resulting solution into the population
(Eqs.~\eqref{eq:mutation_update}--\eqref{eq:update_worst}) &
W (when due; Eq.~\eqref{eq:de_due}) \\
\bottomrule
\end{tabularx}
\end{table}

\subsubsection{Elite Memory Composition and Archive Maintenance}
\label{subsub:elite_maintenance}

The framework maintains the $k$ best unique configurations, with $k\ll N$:
\begin{equation}
    \mathcal{E}_t = \bigl\{\mathbf{x}^*_1, \mathbf{x}^*_2, \dots, \mathbf{x}^*_k\bigr\} \subset \{0,1\}^d\setminus\{\mathbf{0}\},
    \label{eq:elite_archive_set}
\end{equation}
sorted by scalar fitness, $f(\mathbf{x}^*_1) \le \dots \le f(\mathbf{x}^*_k)$. Since $f$ is a single scalar fitness, the archive is a \emph{fitness-ranked top-$k$ set}; no Pareto dominance is defined or used.

Maintenance is performed in Phase~3 of every generation. Let $\mathcal{P}_t=\mathrm{unique}(\mathbf{X}_t^{+}\cup\mathcal{E}_t)$ be the union of the evaluated population and the archive with duplicate bit-strings removed. Then
\begin{equation}
    \mathcal{E}_{t+1} \leftarrow \operatorname{Top-}k\bigl(\mathcal{P}_t;\, f\bigr)
    = \arg\min_{\mathcal{S} \subseteq \mathcal{P}_t,\; |\mathcal{S}|=k} \sum_{\mathbf{x} \in \mathcal{S}} f(\mathbf{x}),
    \label{eq:elite_memory_update}
\end{equation}
i.e., the $k$ lowest-fitness distinct members of $\mathcal{P}_t$. Because $\mathcal{E}_t\subseteq\mathcal{P}_t$, both $f(\mathbf{x}^*_1)$ and $\sum_{\mathbf{x}\in\mathcal{E}}f(\mathbf{x})$ are non-increasing in $t$, so the best-so-far solution is never lost. Diversity within the archive is enforced only through bit-string uniqueness.

\subsubsection{Elite Ordering, Tie-Breaking, Rank-Biased Selection and Feasibility}
\label{subsub:ordering}

\paragraph{Total order and tie-breaking.} Candidates are ranked by the lexicographic key
\begin{equation}
    \bigl(f(\mathbf{x}),\ \|\mathbf{x}\|_1,\ \text{origin},\ \text{index}\bigr),
\end{equation}
all ascending: (i) lower fitness first; (ii) on equal fitness, fewer selected features first; (iii) on equal size, archive members before population members; (iv) otherwise the lower population index. Duplicate bit-strings are removed before ranking, keeping the copy with the best key, so Top-$k$ in Eq.~\eqref{eq:elite_memory_update} is deterministic.

\paragraph{Rank-biased selection.} When Level 2 draws an elite for \ABHG{}, rank $r\in\{1,\dots,k\}$ is chosen with the linear-ranking probability
\begin{equation}
    \Pr[r] = \frac{2\,(k-r+1)}{k\,(k+1)},
    \label{eq:rank_prob}
\end{equation}
which sums to $1$ over $r=1,\dots,k$.

\paragraph{Feasibility.} All-zero vectors are excluded from $\mathcal{P}_t$. A \DE{} trial $\mathbf{u}^{(i)}$ with $\|\mathbf{u}^{(i)}\|_1=0$ is rejected in Eq.~\eqref{eq:greedy_selection_de}, i.e., treated as $f=+\infty$.

\subsubsection{Archive-Level DE Refinement}
\label{subsub:de_refinement}

\paragraph{Schedule.} The refinement period is defined \emph{only} in main-search evaluations, through the interval $m_{\mathrm{FE}}$. Let $n_{\mathrm{DE}}$ be the number of refinement calls so far. At the end of a generation (after Phase~3), a refinement is due when
\begin{equation}
    FE_{\mathrm{main}} \ge (n_{\mathrm{DE}}+1)\,m_{\mathrm{FE}}
    \quad\text{and}\quad FE + k \le FE_{\max}.
    \label{eq:de_due}
\end{equation}
Equivalently, \DE{} is called about every $m_{\mathrm{FE}}/(N-1)$ generations. The second condition guarantees that the budget is never exceeded. The kernel is applied to the archive members only; it neither reads nor modifies $\mathbf{X}$ until the reinsertion step below, and it requires $k\ge3$ so that three distinct indices $i,r,s$ exist. Each call evaluates $k$ trial vectors ($k$ additional evaluations); reinsertion consumes none, since archive fitness values are already known. 
\begin{itemize}
    \item \textbf{Discrete mutation:} For each target $\mathbf{x}^*_i\in\mathcal{E}_{t+1}$, a binary mutant is generated by bitwise operations:
    \begin{equation}
        \mathbf{x}_{\text{mut}}^{(i)} = \mathbf{x}^*_i \oplus \left[ \left( \mathbf{r} < \psi_t \right) \otimes \left( \mathbf{x}^*_r \oplus \mathbf{x}^*_s \right) \right],
        \label{eq:mutation_update}
    \end{equation}
    where $\mathbf{x}^*_r,\mathbf{x}^*_s\in\mathcal{E}_{t+1}$ are archive members with $i,r,s$ pairwise distinct, $\mathbf{r}\in[0,1]^d$ is a uniform random vector (the comparison is element-wise), and $\oplus$, $\otimes$ denote bitwise \textsc{xor} and \textsc{and}. The mask probability $\psi_t=g(\theta_t)$ is the value supplied by the $\mathcal{M}$-Engine at the time of the call.

    \item \textbf{Crossover:} A trial vector $\mathbf{u}^{(i)}$ is formed with crossover rate $C_r\in[0,1]$:
    \begin{equation}
        u^{(i)}_j =
        \begin{cases}
            x_{\text{mut},j}^{(i)}, & \text{if } \text{rand}(0,1) \le C_r \text{ or } j = j_{\text{rand}}, \\
            x_{i,j}^*, & \text{otherwise,}
        \end{cases}
        \label{eq:trial_vector_update}
    \end{equation}
    where $j_{\text{rand}}\in\{1,\dots,d\}$ is a random locus ensuring that at least one mutant bit is inherited.

    \item \textbf{Selection:} The member is replaced only on strict improvement and only if the trial is feasible and not already in the archive:
    \begin{equation}
        \mathbf{x}^*_i \leftarrow
        \begin{cases}
            \mathbf{u}^{(i)}, & \text{if } f(\mathbf{u}^{(i)}) < f(\mathbf{x}^*_i) \text{ and } \mathbf{u}^{(i)}\notin\mathcal{E}_{t+1}, \\
            \mathbf{x}^*_i, & \text{otherwise.}
        \end{cases}
        \label{eq:greedy_selection_de}
    \end{equation}
\end{itemize}

After selection, $\mathcal{E}_{t+1}$ is re-sorted and copied into the population by \emph{elite reinsertion},
\begin{equation}
    \mathbf{X}_{\text{worst}} \leftarrow \mathcal{E}_{t+1},
    \label{eq:update_worst}
\end{equation}
where $\mathbf{X}_{\text{worst}}$ denotes the $k$ lowest-performing individuals of $\mathbf{X}_t^{+}$ (their fitness is replaced by the known archive fitness). Reinsertion is a write into the \emph{population}, not into the archive; it deliberately injects elite structure into Level 2 and thus reduces diversity locally. This is why it is applied only when \DE{} is due and is limited to $k\ll N$ individuals. The resulting diversity change is observed by Level 1 through $DI_t$ in the next generation, which closes the feedback loop.

\subsection{Algorithm Workflow and Architectural Integration}

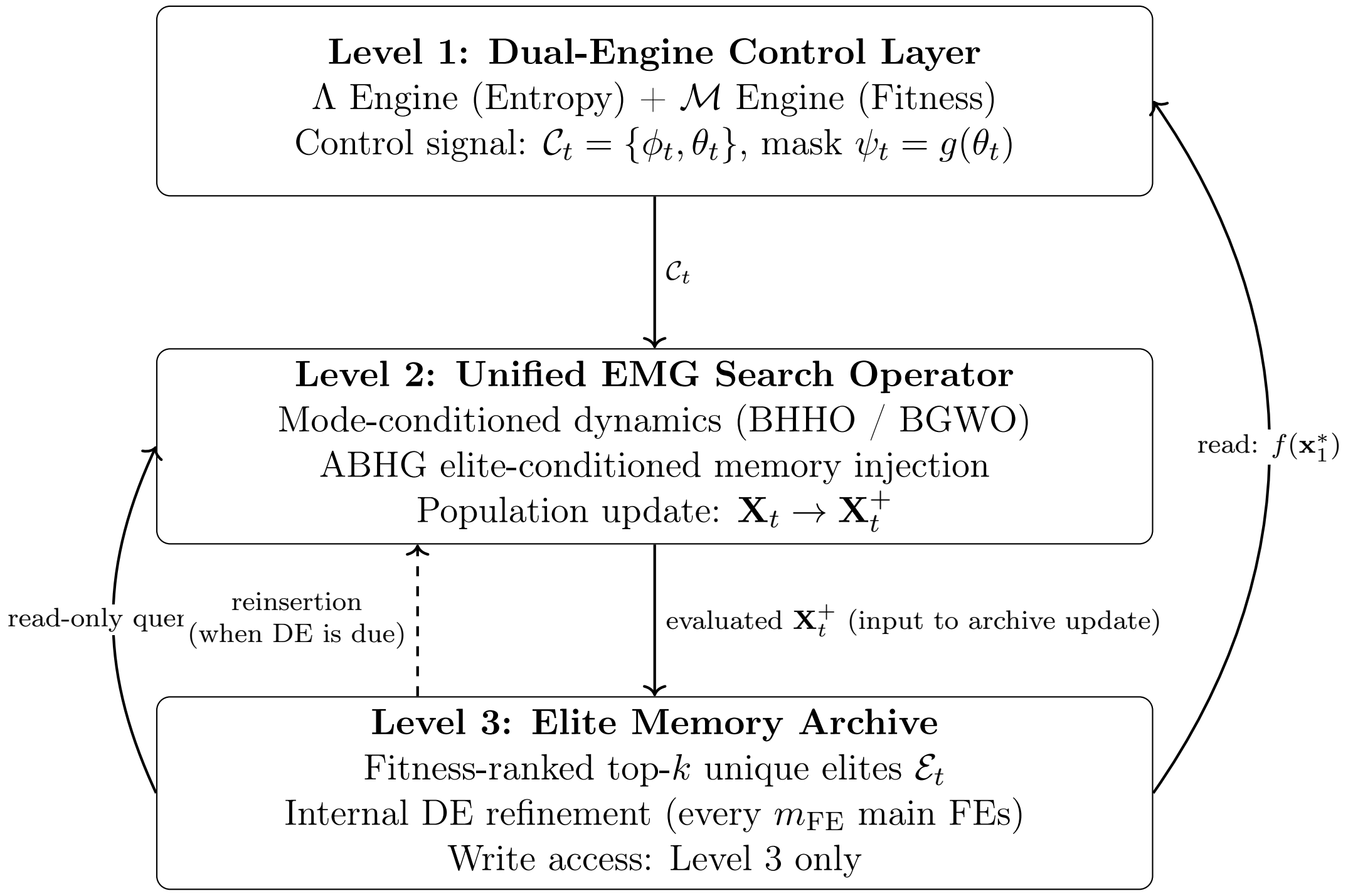
\begin{figure}
\centering
\begin{tikzpicture}[
    box/.style={rectangle, draw, rounded corners, align=center, minimum width=10.5cm, minimum height=2cm},
    arrow/.style={->, thick},
    lbl/.style={font=\scriptsize, align=center, fill=white, inner sep=1pt}
]

\node[box] (l1) {
\textbf{Level 1: Dual-Engine Control Layer} \\
$\Lambda$ Engine (Entropy) + $\mathcal{M}$ Engine (Fitness) \\
Control signal: $\mathcal{C}_t = \{\phi_t, \theta_t\}$, mask $\psi_t=g(\theta_t)$
};

\node[box, below=1.6cm of l1] (l2) {
\textbf{Level 2: Unified EMG Search Operator} \\
Mode-conditioned dynamics (BHHO / BGWO) \\
ABHG elite-conditioned memory injection \\
Population update: $\mathbf{X}_t \rightarrow \mathbf{X}_t^{+}$
};

\node[box, below=1.6cm of l2] (l3) {
\textbf{Level 3: Elite Memory Archive} \\
Fitness-ranked top-$k$ unique elites $\mathcal{E}_t$ \\
Internal DE refinement (every $m_{\mathrm{FE}}$ main FEs) \\
Write access: Level 3 only
};

\draw[arrow] (l1) -- node[lbl, right=2pt] {$\mathcal{C}_t$} (l2);
\draw[arrow] (l2) -- node[lbl, right=2pt] {evaluated $\mathbf{X}_t^{+}$ (input to archive update)} (l3);

\draw[arrow] (l3.east) to[bend right=35] node[lbl] {read: $f(\mathbf{x}^*_1)$} (l1.east);

\draw[arrow] (l3.west) to[bend left=25] node[lbl] {read-only queries} (l2.west);

\draw[arrow, dashed] ([xshift=-2.5cm]l3.north) -- node[lbl, left=2pt] {reinsertion\\(when DE is due)} ([xshift=-2.5cm]l2.south);

\end{tikzpicture}
\caption{Three-level hierarchical architecture of EMG-Hybrid. Level 1 reads the population diversity and the best archived fitness and broadcasts the control channel $\mathcal{C}_t$. Level 2 executes the mode-conditioned operator $\mathcal{O}_{\mathrm{EMG}}$ on $\mathbf{X}_t$, integrating BHHO/BGWO dynamics with ABHG elite injections, and accesses the archive through read-only queries. Level 3 maintains the fitness-ranked top-$k$ archive $\mathcal{E}_t$, which is written only by Level 3 itself (per-generation consolidation and periodic DE refinement), and periodically reinserts refined elites into the population (dashed arrow).}
\label{fig:emg_arch}
\end{figure}

A conceptual overview of the proposed Elite Memory Gravity Hybrid (EMG-Hybrid) is shown in Fig.~\ref{fig:emg_arch}, and the full procedure, including inter-layer interactions, is given in Algorithm~\ref{alg:proposed_framework}.
\begin{algorithm}
\small
\caption{EMG-Hybrid (FE-based budget and DE schedule)}
\label{alg:proposed_framework}
\begin{algorithmic}[1]
\Require $N$, $FE_{\max}$, $k$, $m_{\mathrm{FE}}$, and the parameters of Table~\ref{tab:parameter_settings_ranges}
\Ensure Best subset $\mathbf{x}^*_1\in\mathcal{E}$
\State $\mathbf{X}\leftarrow$ random binary population, repaired; evaluate $f(\mathbf{X})$
\State $FE\leftarrow N$; $FE_{\mathrm{main}}\leftarrow N$; $n_{\mathrm{DE}}\leftarrow0$; $t\leftarrow0$; $t_{\mathrm{last}}\leftarrow0$
\State $\mathcal{E}\leftarrow\mathrm{Top}\text{-}k(\mathbf{X};f)$; $\phi\leftarrow\BHHO$; $\theta\leftarrow\theta_0$; $s\leftarrow0$; $f^*_{\mathrm{prev}}\leftarrow\infty$
\While{$FE+N-1\le FE_{\max}$}
  \State $t\leftarrow t+1$; $\mathbf{X}^{\mathrm{old}}\leftarrow\mathbf{X}$; $\omega\leftarrow FE/FE_{\max}$
  \Statex \hspace{\algorithmicindent}\textit{// Level 1: decision layer}
  \State Compute the normalized diversity index $DI_t$ from $\mathbf{X}$ using Eq.~\eqref{eq:diversity_index}
  \If{$DI_t\le\tau$ \textbf{and} $t-t_{\mathrm{last}}\ge t_{\min}$}
    \State $\phi\leftarrow\overline{\phi}$; $t_{\mathrm{last}}\leftarrow t$ \Comment{$\Lambda$-engine, Eq.~\eqref{eq:switching_rule}}
  \EndIf
  \State $\Delta\leftarrow f^*_{\mathrm{prev}}-f(\mathbf{x}^*_1)$; $f^*_{\mathrm{prev}}\leftarrow f(\mathbf{x}^*_1)$ \Comment{Eq.~\eqref{eq:stagnation_delta}; $\Delta=\infty$ at $t=1$}
  \State $\hat{s}\leftarrow s+1$ if $\Delta\le\epsilon$, else $\hat{s}\leftarrow0$ \Comment{Eq.~\eqref{eq:stagn_counter}}
  \State Update $\theta$ by Eq.~\eqref{eq:theta_update} using $\hat{s}$; $s\leftarrow0$ if $\hat{s}\ge S_{\mathrm{tol}}$, else $s\leftarrow\hat{s}$ \Comment{$\mathcal{M}$-engine, Eq.~\eqref{eq:stagn_reset}}
  \Statex \hspace{\algorithmicindent}\textit{// Level 2: unified search (archive read-only)}
  \State $\mathbf{V}\leftarrow$ \textbf{if} $\phi=\BHHO$ \textbf{then} Eq.~\eqref{eq:bhho_update} \textbf{else} Eq.~\eqref{eq:bgwo_update} (with $a=2(1-\omega)$)
  \State $\mathbf{X}\leftarrow\mathcal{T}(\mathbf{V})$ \Comment{clip, sigmoid, Bernoulli; Eqs.~\eqref{eq:transfer}--\eqref{eq:bernoulli_bit}}
  \State $\mathbf{X}\leftarrow\ABHG(\mathbf{X},\mathcal{E})$ \Comment{Eqs.~\eqref{eq:abhg_probability}--\eqref{eq:abhg_bit_rule}}
  \State Flip each bit with probability $1/d$; repair empty subsets \Comment{Eq.~\eqref{eq:repair}}
  \State $\mathbf{x}_N\leftarrow\mathbf{x}^*_1$ with its known fitness \Comment{elite clone, after mutation and repair}
  \State Evaluate $f(\mathbf{x}_i)$ for $i<N$; $FE\leftarrow FE+N-1$; $FE_{\mathrm{main}}\leftarrow FE_{\mathrm{main}}+N-1$
  \State $f_{\mathrm{elite}}\leftarrow f(\mathbf{x}^*_1)$ \Comment{best fitness of the current archive $\mathcal{E}$, not yet updated}
  \State For $i<N$: restore $\mathbf{x}^{\mathrm{old}}_i$ if $f(\mathbf{x}_i)>f(\mathbf{x}^{\mathrm{old}}_i)+\theta\,\max\!\left(f_{\mathrm{elite}},10^{-3}\right)$ \Comment{Eq.~\eqref{eq:theta_acceptance_l2}; clone exempt}
  \Statex \hspace{\algorithmicindent}\textit{// Level 3: archive update, then DE refinement if due}
  \State $\mathcal{E}\leftarrow\mathrm{Top}\text{-}k(\mathrm{unique}(\mathbf{X}\cup\mathcal{E});f)$ \Comment{Eq.~\eqref{eq:elite_memory_update}}
  \If{$FE_{\mathrm{main}}\ge(n_{\mathrm{DE}}+1)\,m_{\mathrm{FE}}$ \textbf{and} $FE+k\le FE_{\max}$}
    \State $\psi\leftarrow g(\theta)$; $\mathcal{E}\leftarrow\textsc{DE-Refine}(\mathcal{E},\psi)$ \Comment{Eqs.~\eqref{eq:mutation_update}--\eqref{eq:greedy_selection_de}}
    \State $n_{\mathrm{DE}}\leftarrow n_{\mathrm{DE}}+1$; $FE\leftarrow FE+k$
    \State Replace the $k$ worst individuals of $\mathbf{X}$ by $\mathcal{E}$ \Comment{Eq.~\eqref{eq:update_worst}, no evaluations}
  \EndIf
\EndWhile
\State \Return $\mathbf{x}^*_1(\mathcal{E})$
\end{algorithmic}
\end{algorithm}

\subsection{Time Complexity and Evaluation Budget}
\label{subsec:time_complexity}

The cost of EMG-Hybrid is analyzed in terms of fitness evaluations (FE), which dominate the runtime of wrapper-based feature selection. Let
\begin{equation*}
\begin{aligned}
N &= \text{population size}, &
k &= \text{archive size},\\
d &= \text{dimensionality of the binary space}, &
F &= \text{cost of one wrapper fitness evaluation},\\
FE_{\max} &= \text{total evaluation budget}, &
FE_{\mathrm{main}} &= \text{main-search evaluations (including initialization)},\\
m_{\mathrm{FE}} &= \text{DE interval in main-search FEs}, &
n_{\mathrm{DE}} &= \text{number of DE calls}.
\end{aligned}
\end{equation*}
$F$ is the cost of training and validating the classifier on a subset. It grows with the number of training samples and with the number of selected features, and it is the dominant term in practice.

\paragraph{Evaluation budget and overhead.} The initial population costs $N$ evaluations and each generation costs $N-1$, because the elite clone keeps its known fitness. DE evaluations are counted separately from the main search, so the total number of evaluations is
\begin{equation}
    FE = FE_{\mathrm{main}} + k\,n_{\mathrm{DE}} \le FE_{\max}, \qquad n_{\mathrm{DE}} \le \Bigl\lfloor \frac{FE_{\mathrm{main}}}{m_{\mathrm{FE}}} \Bigr\rfloor ,
    \label{eq:nfe}
\end{equation}
and the number of main-search generations is $(FE_{\mathrm{main}}-N)/(N-1)=O(FE_{\mathrm{main}}/N)$. The relative evaluation overhead of the refinement is
\begin{equation}
    \varrho = \frac{k\,n_{\mathrm{DE}}}{FE_{\mathrm{main}}} \le \frac{k}{m_{\mathrm{FE}}} = \frac{5}{200} = 2.5\,\% .
    \label{eq:overhead}
\end{equation}
This bound is reached only if all scheduled refinements are completed. For $N=10$ and $FE_{\max}=2000$, at most $9$ DE calls fit into the budget ($45$ evaluations), so $\varrho\approx2.3\,\%$ and a run consists of about $215$ generations. Because the budget $FE_{\max}$ is shared, the DE evaluations do not add to the total of an equal-budget comparison; instead, they reduce the number of main-search generations by about $\varrho$.

\paragraph{Cost per component.} Table~\ref{tab:de_cost} lists the cost of the individual components. All search and bookkeeping operations are $O(Nd)$ or lower, so they are negligible compared with the $(N-1)F$ cost of the wrapper evaluations.

\begin{table}[t]
\centering
\caption{Cost of EMG-Hybrid components ($F$: cost of one fitness evaluation).}
\label{tab:de_cost}
\small
\begin{tabular}{@{}lll@{}}
\toprule
Component & Cost & Frequency \\
\midrule
Level 1: $DI_t$, $\Delta_t$, $\theta_t$, $\phi_t$ & $O(Nd)$ & every generation \\
Level 2: $\mathcal{U}_{\phi_t}$, $\mathcal{T}$, mutation, repair & $O(Nd)$ (\BGWO{}: $\approx 3Nd$) & every generation \\
Level 2: \ABHG{} (Hamming distance and bit copying) & $O(p_{\mathrm{abhg}}Nd)$ & every generation \\
Fitness evaluation of the population & $(N-1)\,F$ & every generation \\
Level 3: archive update (unique + Top-$k$) & $O(Nd + N\log N)$ & every generation \\
Level 3: \DE{} refinement & $O(kd) + k\,F$ & when due (Eq.~\eqref{eq:de_due}) \\
Reinsertion (synchronization) & $O(kd + N\log N)$, no evaluations & when \DE{} is due \\
\bottomrule
\end{tabular}
\end{table}

\paragraph{Main search.} One generation consists of the population update, the diversity computation, the \ABHG{} operations (each $O(Nd)$), archive maintenance ($O(Nd+N\log N)$ with hashing-based duplicate checking), and the wrapper evaluations ($O(NF)$). Hence
\begin{equation}
C_{\mathrm{main,gen}} = O\bigl(N(F+d+\log N)\bigr),
\qquad
C_{\mathrm{main}} = O\Bigl(\tfrac{FE_{\mathrm{main}}}{N}\Bigr) C_{\mathrm{main,gen}} = O\bigl(FE_{\mathrm{main}}(F+d+\log N)\bigr).
\label{eq:main_generation_complexity}
\end{equation}

\paragraph{DE refinement.} A DE call performs mutation, crossover and repair in $O(kd)$ and $k$ wrapper evaluations in $O(kF)$. Selection, the duplicate check against the archive, and reinsertion do not change this bound with hashing-based uniqueness checking. With the call bound of Eq.~\eqref{eq:nfe},
\begin{equation}
C_{\mathrm{DE,call}} = O\bigl(k(F+d+\log k)\bigr),
\qquad
C_{\mathrm{DE}} = n_{\mathrm{DE}}\,C_{\mathrm{DE,call}} = O\Bigl(\frac{FE_{\mathrm{main}}}{m_{\mathrm{FE}}}\,k(F+d+\log k)\Bigr).
\label{eq:de_total_complexity}
\end{equation}

\paragraph{Total.} Combining both parts gives
\begin{equation}
C_{\mathrm{total}} = O\Bigl(FE_{\mathrm{main}}(F+d+\log N) + \frac{FE_{\mathrm{main}}}{m_{\mathrm{FE}}}\,k(F+d+\log k)\Bigr).
\label{eq:total_complexity_fe}
\end{equation}
For high-dimensional feature selection, where $d$ dominates the logarithmic terms, this simplifies to
\begin{equation}
C_{\mathrm{total}} = O\Bigl(FE_{\mathrm{main}}(F+d)\Bigl(1+\frac{k}{m_{\mathrm{FE}}}\Bigr)\Bigr).
\label{eq:total_complexity_simplified}
\end{equation}
For fixed $k$ and $m_{\mathrm{FE}}$, the refinement is a constant-factor overhead of at most $k/m_{\mathrm{FE}}=2.5\%$, so the overall complexity remains
\begin{equation}
C_{\mathrm{total}} = O\bigl(FE_{\mathrm{main}}(F+d)\bigr).
\label{eq:final_complexity}
\end{equation}
Thus archive-level DE refinement preserves the asymptotic order of the underlying population-based search and adds only a bounded evaluation and processing overhead. Measured wall-clock times may deviate from this evaluation-based analysis, because $F$ itself depends on the size of the evaluated subsets.

\section{Experiments and Results}\label{sec5}
Fourteen benchmark datasets from the UCI repository and OpenML, spanning low- to high-dimensional feature spaces, were employed to evaluate the performance of the proposed hybrid adaptive metaheuristic in feature selection. Experiments were conducted in Python on a system with an Intel Core i7 (2.4 GHz) processor, 16 GB of RAM, and Windows 11, using 3-fold cross-validation. Performance was assessed in terms of fitness, compactness of selected feature subsets, and classification accuracy, with dataset details summarized in \autoref{tab1}.

\subsection{Experimental Settings and Leakage-Free Evaluation Protocol}
\label{subsec:experimental_protocol}

\paragraph{Nested cross-validation.} All experiments employ a leakage-free nested cross-validation (CV) framework with stratified three-fold CV in both the outer and the inner loops. For each outer fold, all preprocessing operations are fitted exclusively on the corresponding training data. Numerical features are mean-imputed and standardized, whereas categorical features are imputed with a dedicated \texttt{missing} category and one-hot encoded; unseen categories are handled with \texttt{handle\_unknown="ignore"}. Within each inner fold, the preprocessing pipeline is fitted on the inner-training partition and applied to the corresponding validation partition. After feature selection, the preprocessing pipeline and the classifier are fitted on the complete outer-training partition and evaluated on the outer-test partition. This procedure separates feature selection and model selection from the final performance estimate and thereby reduces selection bias~\cite{varma2006bias}.

\paragraph{Objective function.} Feature selection and hyperparameter tuning use the outer-training data only. For a candidate subset $\mathbf{x}$, the objective is
\begin{equation}
f(\mathbf{x}) = \eta\,\overline{E}_{\mathrm{inner}}(\mathbf{x}) + \mu\,\frac{|\mathbf{x}|}{d},
\label{eq:inner_fitness}
\end{equation}
where $\overline{E}_{\mathrm{inner}}$ is the mean $k$-NN classification error ($k=5$) over the inner validation folds, $|\mathbf{x}|$ is the number of selected features, and $d$ is the number of encoded features. The weights are fixed at $\eta=0.99$ and $\mu=0.01$ for all methods, and all methods use the same inner-fold splits and the same objective within each outer fold.

\paragraph{Hyperparameter tuning.} Each method is tuned by random search with the same budget of $B=15$ candidate configurations per outer fold. Candidates are sampled uniformly at random from the tuning ranges listed in Table~\ref{tab:parameter_settings_ranges}, and each candidate is scored by the mean inner-fold fitness of the subset found by the optimizer. The configuration with the lowest score is selected for that outer fold; hence the selected values may differ across outer folds, and Table~\ref{tab:parameter_settings_ranges} should not be read as a single fold-independent set of tuned values. Its \emph{Value} column gives the reference configuration, and its \emph{Range} column gives the interval used for sampling and for the sensitivity analysis (Section~\ref{sec:sensitivity}). Three groups of parameters are not tuned: (i) quantities fixed by the protocol, namely the population size $N=10$ of the population-based methods, the budget $FE_{\max}=2000$, and the weights $\eta,\mu$, since candidates with different objective weights would not be comparable; (ii) structural constants marked as fixed in the table (for example the number of leaders and the transfer functions of EHHO and MBHO); and (iii) parameters without a range. For EMG-Hybrid, sampled configurations that violate $\theta_{\min}\le\theta_0\le\theta_{\max}$, $\psi_{\min}<\psi_{\max}$, or $\lambda_{\mathrm{stagn}}>\lambda_{\mathrm{up}}>1>\lambda_{\mathrm{down}}$ are rejected and resampled.

\paragraph{Evaluation budget.} To compare the main optimization procedures fairly, all methods use an identical budget of $FE_{\max}=2000$ objective-function evaluations per run. The budget includes the initial population and every later evaluation, including those of refinement operators such as the DE kernel of EMG-Hybrid; the number of generations is therefore adjusted to the evaluation cost of each algorithm. The tuning evaluations are \emph{not} part of this budget. Every method receives the same number of candidates ($B=15$), but the number of tunable parameters differs: a method with many tunable parameters, such as EMG-Hybrid, covers its parameter space less thoroughly with 15 candidates than a method with two or three. The equal-budget statement therefore applies to the main optimization stage only and does not imply equal total computational cost including tuning.

\paragraph{Final evaluation and reporting.} After tuning, the selected configuration of each outer fold is used for ten independent stochastic runs, with the same predefined seeds across competing methods. In each run, the selected subset is used to fit the $k$-NN classifier on the complete outer-training partition, and its accuracy is measured once on the unseen outer-test partition. The value reported for each dataset is the mean over the ten runs and the three outer folds. No outer-test information is used in preprocessing, feature selection, hyperparameter tuning, or archive updates. Three quantities are reported separately: inner-fold wrapper fitness (Table~\ref{tab9}), outer-fold test accuracy (Table~\ref{tab7}), and the number of selected features (Table~\ref{tab8}).

\subsection{Parameter Sensitivity Analysis}
\label{sec:sensitivity}

A sensitivity analysis was conducted to examine the effects of selected EMG-Hybrid parameters on optimization performance. The reference configuration corresponds to the values listed in the \emph{Value} column of Table~\ref{tab:parameter_settings_ranges} and is labelled ``Default'' in Tables~\ref{tab-stagnation}--\ref{tab:sensitivity-rhomax}. For each parameter, two alternative values within its specified tuning range were evaluated, one below and one above the reference value, while all remaining parameters were fixed. This analysis was performed independently of the hyperparameter tuning procedure, without additional random search. All configurations used the same population size, evaluation budget ($FE_{\max}=2000$), number of independent runs, and cross-validation protocol as the main experiments. Evaluation was conducted on five datasets: Darwin, Toxicity, Micro-mass, Leukemia, and GCM.

Performance was assessed using mean classification accuracy, mean number of selected features, and mean best fitness. Higher accuracy and lower feature counts and fitness values indicate better performance. Since different configurations were run independently and were not necessarily initialized with identical random seeds, the results are interpreted descriptively rather than through paired statistical tests. Furthermore, the reference values were selected through the tuning procedure and may therefore have an advantage over the alternative settings. The observed differences, which are generally small and mostly below $0.02$ in accuracy, are consequently used to characterize parameter sensitivity rather than to establish statistically optimal values.

\paragraph{Stagnation patience $S_{\mathrm{tol}}$.}
Table~\ref{tab-stagnation} examines $S_{\mathrm{tol}}$, the number of consecutive generations without a meaningful improvement of the best archived fitness ($\Delta_t\le\epsilon$) after which stagnation is declared. When stagnation is declared, the acceptance threshold is multiplied by $\lambda_{\mathrm{stagn}}$ (clipped at $\theta_{\max}$) and the stagnation counter is reset. $S_{\mathrm{tol}}$ therefore controls \emph{when} the threshold is expanded, whereas $\lambda_{\mathrm{stagn}}$ controls \emph{how much} it is expanded. A smaller $S_{\mathrm{tol}}$ triggers expansions sooner, which can help the search escape local optima but also admits inferior candidates more often. A larger $S_{\mathrm{tol}}$ delays expansions and keeps the acceptance rule stricter for longer, at the risk of a prolonged stagnation. The results show no consistent trend: the smaller value was generally worse than the default, the larger value was better on some datasets and worse on others, and the default was the most favourable setting on most of the datasets in accuracy and fitness. Since the runs of different configurations were not paired by seed, the larger differences observed on some high-dimensional datasets cannot be attributed to $S_{\mathrm{tol}}$ with certainty. Overall, no value dominates on all datasets, and the default is a reasonable compromise rather than a uniform optimum.

\paragraph{Diversity threshold $\tau$.}
Table~\ref{tab:sensitivity-tau} examines the effect of the entropy threshold used by the $\Lambda$-Engine. A phase switch between \BHHO{} and \BGWO{} is triggered when $DI_t\leq\tau$ and the minimum phase inertia $t_{\min}$ has elapsed. A larger $\tau$ makes this condition easier to satisfy, allowing phase switching at higher diversity levels, whereas a smaller $\tau$ requires a greater reduction in diversity before switching and may prolong the current search phase. Across all five datasets, $\tau=0.6$ produced the smallest average feature subsets. The default setting achieved the highest accuracy and lowest average best fitness on four datasets (Darwin, toxicity, micro-mass, and GCM), while $\tau=0.6$ yielded marginally better results on leukemia for both criteria. The $\tau=0.4$ setting was not optimal in any of the three reported criteria for any dataset. Overall, the results indicate that increasing $\tau$ favours subset compactness, while the default threshold provides a more favourable accuracy--fitness balance on most of the evaluated datasets. The observed accuracy differences were relatively small.

\paragraph{ABHG application rate $p_{\mathrm{abhg}}$.}
Table~\ref{tab:sensitivity-pabhg} examines $p_{\mathrm{abhg}}$, the probability that an individual is selected for the ABHG assimilation step in a generation. A higher value applies elite bit copying to more individuals, which strengthens exploitation but may reduce diversity. A lower value leaves more individuals unchanged by the elite memory, so their structure is preserved, but the transfer of elite information is slower. The default value was the most favourable setting in accuracy and fitness on most datasets, and the larger value gave the most compact subsets on all datasets. As for $\tau$, the differences in accuracy were small, and the subset size decreased steadily as the parameter increased. These observations are consistent with a stronger elite influence producing more compact, but not more accurate, subsets. Diversity was not measured directly in this analysis, so a loss of diversity is a plausible explanation rather than an established effect.

\paragraph{Maximum Interaction Radius $\rho_{\max}$.}
Table~\ref{tab:sensitivity-rhomax} examines $\rho_{\max}$, the upper bound of the per-bit assimilation probability $\rho_i=\rho_{\max}\Gamma_i$ with $\Gamma_i\in(0,1]$ (Eqs.~\eqref{eq:abhg_probability}--\eqref{eq:abhg_rate}). In an ABHG update only bits that differ from the selected elite can be copied, and each is copied with probability $\rho_i\le\rho_{\max}$, so $\rho_{\max}$ limits the expected fraction of differing bits that one update can transfer. A larger value permits larger changes toward the elite, whereas a smaller value gives more conservative updates that preserve more of the candidate's structure and slow the assimilation. The default value was the most favourable setting in accuracy and fitness on most datasets, and the larger value gave the most compact subsets on all datasets. As for the other ABHG parameter, the subset size decreased steadily as $\rho_{\max}$ increased, while the accuracy was highest at or near the default value, with small differences.

\begin{table}[ht]
\renewcommand{\arraystretch}{1} 
\caption{Performance Comparison for Different Stagnation Tolerance $S_{\text{tol}}$ Values}\label{tab-stagnation}
\begin{tabularx}{\textwidth}{l X X X X X X X X X}
\toprule
\textbf{Dataset} & \multicolumn{3}{c}{\textbf{Average Accuracy}} & \multicolumn{3}{c}{\textbf{Average Feature Selection}} & \multicolumn{3}{c}{\textbf{Average Best Fitness}} \\
\cmidrule(lr){2-4} \cmidrule(lr){5-7} \cmidrule(lr){8-10}
 & 3.0 & Default  & 7.0 &3.0 & Default & 7.0 & 3.0 & Default & 7.0 \\
\midrule

darwin & 0.941& 0.969 &\textbf{0.97} &49.07 &\textbf{46.7}&54.17 &0.0281 &\textbf{0.0269}&0.0258\\
toxicity& 0.832& \textbf{0.843}&0.838 &132.8 &131.74&\textbf{128.07} &0.0781 &\textbf{0.0774}&0.0776\\
micro-mass &0.859 & \textbf{0.87}  &0.865 &193.53 &\textbf{182.43}&198.7 &0.102 &\textbf{0.0796}&0.106 \\
leukemia &\textbf{0.92} & 0.937&0.918 &1429.07 &1271.87&\textbf{969.47} &0.093 &0.087&\textbf{0.076}\\
GCM &0.691 & \textbf{0.709}&0.684 & 954.67&\textbf{943.01}& 962.63& 0.1554&\textbf{0.1334}&0.1643\\
\bottomrule
\end{tabularx}
\end{table}

\begin{table}[ht]
\centering
\renewcommand{\arraystretch}{1.1}
\caption{Sensitivity Analysis of the Diversity Threshold $\tau$}
\label{tab:sensitivity-tau}
\begin{tabularx}{\textwidth}{l *{9}{>{\centering\arraybackslash}X}}
\toprule
\textbf{Dataset} &
\multicolumn{3}{c}{\textbf{Average Accuracy}} &
\multicolumn{3}{c}{\textbf{Average Feature Selection}} &
\multicolumn{3}{c}{\textbf{Average Best Fitness}} \\
\cmidrule(lr){2-4}
\cmidrule(lr){5-7}
\cmidrule(lr){8-10}
& 0.4 & Default & 0.6
& 0.4 & Default & 0.6
& 0.4 & Default & 0.6 \\
\midrule
darwin
& 0.965 & \textbf{0.969} & 0.968
& 48.1 & 46.7 & \textbf{45.3}
& 0.0308 & \textbf{0.0269} & 0.0275 \\

toxicity
& 0.839 & \textbf{0.843} & 0.841
& 134.4 & 131.7 & \textbf{129.1}
& 0.0810 & \textbf{0.0774} & 0.0792 \\

micro-mass
& 0.864 & \textbf{0.870} & 0.867
& 187.9 & 182.4 & \textbf{177.0}
& 0.0854 & \textbf{0.0796} & 0.0819 \\

leukemia
& 0.915 & 0.937 & \textbf{0.918}
& 1322.7 & 1271.9 & \textbf{1221.0}
& 0.0895 & 0.0870 & \textbf{0.0854} \\

GCM
& 0.701 & \textbf{0.709} & 0.705
& 961.9 & 943.0 & \textbf{933.6}
& 0.1407 & \textbf{0.1334} & 0.1369 \\
\bottomrule
\end{tabularx}
\end{table}

\begin{table}[ht]
\centering
\renewcommand{\arraystretch}{1.1}
\caption{Sensitivity Analysis of the ABHG Parameter $p_{\mathrm{ABHG}}$}
\label{tab:sensitivity-pabhg}
\begin{tabularx}{\textwidth}{l *{9}{>{\centering\arraybackslash}X}}
\toprule
\textbf{Dataset} &
\multicolumn{3}{c}{\textbf{Average Accuracy}} &
\multicolumn{3}{c}{\textbf{Average Feature Selection}} &
\multicolumn{3}{c}{\textbf{Average Best Fitness}} \\
\cmidrule(lr){2-4}
\cmidrule(lr){5-7}
\cmidrule(lr){8-10}
& 0.1 & Default & 0.3
& 0.1 & Default & 0.3
& 0.1 & Default & 0.3 \\
\midrule
darwin
& 0.967 & \textbf{0.969} & 0.968
& 47.6 & 46.7 & \textbf{45.8}
& 0.0289 & \textbf{0.0269} & 0.0276 \\

toxicity
& 0.841 & \textbf{0.843} & 0.842
& 134.4 & 131.7 & \textbf{129.1}
& 0.0792 & \textbf{0.0774} & 0.0783 \\

micro-mass
& 0.867 & \textbf{0.870} & 0.868
& 187.9 & 182.4 & \textbf{177.0}
& 0.0827 & \textbf{0.0796} & 0.0810 \\

leukemia
& 0.916 & 0.937 & \textbf{0.918}
& 1310.0 & 1271.9 & \textbf{1233.7}
& 0.0884 & 0.0870 & \textbf{0.0856} \\

GCM
& 0.705 & \textbf{0.709} & 0.706
& 961.9 & 943.0 & \textbf{933.6}
& 0.1371 & \textbf{0.1334} & 0.1360 \\
\bottomrule
\end{tabularx}
\end{table}

\begin{table}[ht]
\centering
\renewcommand{\arraystretch}{1.1}
\caption{Sensitivity Analysis of the Maximum Interaction Radius $\rho_{\max}$}
\label{tab:sensitivity-rhomax}
\begin{tabularx}{\textwidth}{l *{9}{>{\centering\arraybackslash}X}}
\toprule
\textbf{Dataset} &
\multicolumn{3}{c}{\textbf{Average Accuracy}} &
\multicolumn{3}{c}{\textbf{Average Feature Selection}} &
\multicolumn{3}{c}{\textbf{Average Best Fitness}} \\
\cmidrule(lr){2-4}
\cmidrule(lr){5-7}
\cmidrule(lr){8-10}
& 0.05 & Default & 0.2
& 0.05 & Default & 0.2
& 0.05 & Default & 0.2 \\
\midrule
darwin
& 0.966 & \textbf{0.969} & 0.967
& 48.1 & 46.7 & \textbf{44.8}
& 0.0299 & \textbf{0.0269} & 0.0283 \\

toxicity
& 0.840 & \textbf{0.843} & 0.841
& 135.7 & 131.7 & \textbf{127.8}
& 0.0802 & \textbf{0.0774} & 0.0791 \\

micro-mass
& 0.866 & \textbf{0.870} & 0.866
& 189.7 & 182.4 & \textbf{175.1}
& 0.0838 & \textbf{0.0796} & 0.0826 \\

leukemia
& 0.915 & 0.937 & \textbf{0.917}
& 1335.5 & 1271.9 & \textbf{1208.3}
& 0.0897 & 0.0870 & \textbf{0.0861} \\

GCM
& 0.704 & \textbf{0.709} & 0.703
& 971.3 & 943.0 & \textbf{924.1}
& 0.1381 & \textbf{0.1334} & 0.1387 \\
\bottomrule
\end{tabularx}
\end{table}

\begin{table}[ht]
\renewcommand{\arraystretch}{1} 
\caption{Datasets Information for 3-fold cross validation}
\begin{tabular*}{\textwidth}{@{\extracolsep\fill}lcccc}
\toprule
Dataset & No. Attributes & No. Instances & Dimensionality & Noisy \\
\midrule
breast cancer original & 9 & 699 & low & No \\
wine & 13 & 178 & low & No \\
statlog heart & 13 & 270 & low & No \\
zoo & 16 & 101 & low & No \\
vehicle & 18 & 946 & low & Medium \\
waveform & 21 & 5000 & low & Yes \\

HCV & 29 & 1385 & moderate & Medium \\
breast cancer diagnostic & 30 & 569 & moderate & No \\
Ionosphere & 34 & 351 & moderate & Yes \\
darwin & 451 & 174 & moderate & Medium \\

toxicity & 1203 & 171 & high & Yes \\
micro mass & 1300 & 931 & high & Yes \\
leukemia & 7129 & 72 & high & Yes \\
GCM & 16064 & 190 & high & Yes \\
\bottomrule
\end{tabular*}
\label{tab1}
\end{table}

\begin{table*}[t]
\centering
\caption{Parameters of EMG-Hybrid and Baseline Algorithms (with plausible ranges)}
\label{tab:parameter_settings_ranges}
\small
\resizebox{\textwidth}{!}{%
\begin{tabular}{@{}llll@{}}
\toprule
\textbf{Symbol / Parameter} & \textbf{Meaning} & \textbf{Value} & \textbf{Plausible range} \\
\midrule
\multicolumn{4}{l}{\textit{EMG-Hybrid Framework}} \\
\midrule
$N$ & population size & $10$ & $[10,30]$ \\
$FE$ & function evaluations (fixed by protocol) & $2000$ & $[1000,20000]$ \\
$k$ & archive size & $5$ & $[3,10]$ (need $\ge 3$ leaders/DE parents) \\
$\eta,\ \mu$ & fitness weights, $\eta+\mu=1$ & $0.99$, $0.01$ & $\eta\in[0.8,0.99]$, $\mu=1-\eta$ \\
$\tau$ & entropy floor & $0.5$ & $[0.3,0.7]$ \\
$t_{\min}$ & minimum phase inertia (iterations) & $10$ & $[5,20]$ \\
$\epsilon$ & stagnation tolerance & $10^{-4}$ & $[10^{-6},10^{-3}]$ \\
$S_{\mathrm{tol}}$ & stagnation patience & $5$ & $[3,10]$ \\
$\theta_0$ & initial acceptance threshold & $0.01$ & $[0.005,0.02]$ \\
$\theta_{\min}$ & lower threshold bound & $0.001$ & $[0.0005,0.005]$ \\
$\theta_{\max}$ & Upper relative acceptance threshold & $0.3$ & $[0.15,0.6]$ \\
$\lambda_{\mathrm{stagn}}$ & multiplier on stagnation ($>1$) & $1.05$ & $[1.05,1.5]$ \\
$\lambda_{\mathrm{up}}$ & multiplier in exploration ($>1$) & $1.02$ & $[1.01,1.10]$ \\
$\lambda_{\mathrm{down}}$ & multiplier in exploitation ($<1$) & $0.98$ & $[0.90,0.99]$ \\
$\psi_{\min}$ & lower DE mask rate & $0.05$ & $[0.01,0.10]$ \\
$\psi_{\max}$ & upper DE mask rate & $0.5$ & $[0.3,0.7]$ \\
$m_{\mathrm{FE}}$ & DE period (function evaluations) & $200$ & $[5\%,20\%]$ of $FE$ $\Rightarrow [100,400]$ \\
$C_r$ & DE crossover rate & $0.9$ & $[0.5,0.95]$ \\
$\kappa$ & transfer steepness & $1.0$ & $[0.5,2.0]$ \\
$V_{\max}$ & step clipping bound & $4.0$ & $[2,6]$ (bit probability stays in $[\sigma(-V_{\max}),\sigma(V_{\max})]$) \\
$p_{\mathrm{abhg}}$ & ABHG application rate & $0.2$ & $[0.05,0.4]$ \\
$\rho_{\max}$ & ABHG max per-bit rate & $0.1$ & $[0.02,0.2]$ \\
$p_{\mathrm{mut}}$ & bit-flip mutation probability & $1/d$ & $[0.5/d,\,2/d]$ \\
\midrule
\multicolumn{4}{l}{\textit{Baseline Algorithms}} \\
\midrule
PSO Inertia Weights & maximum, minimum inertia weight & $0.9$, $0.4$ & $w_{\max}\in[0.8,0.95]$, $w_{\min}\in[0.2,0.5]$ \\
PSO Acceleration Coefficients & $c_1$, $c_2$ & $2.0$, $2.0$ & $[1.5,2.5]$ each ($c_1+c_2\lesssim 4.5$) \\
GA Crossover / Mutation Rates & crossover rate, mutation rate & $0.8$, $0.2$ & $[0.6,0.95]$, $[0.01,0.2]$ \\
ACO Parameters & initial pheromone, evaporation rate, $Q$ & $0.5$, $0.15$, $1.0$ & $[0.1,1.0]$, $[0.05,0.5]$, $[0.5,10]$ \\
HHO Patience & stagnation patience & $5$ & $[3,10]$ \\
GWO Patience & stagnation patience & $2$ & $[2,10]$ \\
EHHO & population size, iterations, search bounds & $10$, $100$, $[-1,1]$ & $[10,30]$, budget-matched to $FE$, $[-1,1]$ to $[-4,4]$ \\
EHHO & transfer function & $|\tanh(x)|$ & fixed (S/V-shaped family) \\
EHHO & L\'evy exponent, L\'evy scale, leader count & $1.5$, $0.01$, $3$ & $[1.0,2.0]$, $[0.001,0.1]$, $3$ (fixed) \\
MBHO & population size, distinct-mask evaluation budget & $10$, $2000$ & $[10,30]$, fixed by protocol \\
MBHO & mutation rate per star & $0.5$ & $[0.1,0.9]$ \\
MBHO & bit-flip rate, chaotic-init iterations & $1/d$, $10$ & $[0.5/d,\,2/d]$, $[5,50]$ \\
MBHO & transfer function, stall limit, inner CV folds & $|\tan(x)|\ge 0.5$, $100$, $3$ & fixed, $[50,500]$, $[3,5]$ \\
\bottomrule
\end{tabular}%
}
\end{table*}

\subsection{Comparison with Existing Metaheuristics and Forward Selection}
\label{subsec:comparison}

EMG-Hybrid was compared with eight existing metaheuristic feature
selection methods, namely MBHO, EHHO, BGWOHHO, HHO, GWO, PSO, GA,
and ACO, across 14 benchmark datasets. Three performance criteria
were considered: classification accuracy, the number of selected
features, and optimization fitness. For statistical evaluation,
each dataset was treated as a block, and methods were ranked
independently within each dataset. Average ranks were assigned to
tied results. Higher accuracy and lower feature counts and fitness
values were considered preferable. The Friedman test was used to
examine overall differences among the nine methods, followed by
two-sided paired Wilcoxon signed-rank tests comparing EMG-Hybrid
with each baseline. The resulting pairwise $p$-values were
adjusted using the Holm procedure, with a significance level of
$\alpha=0.05$. Forward Selection was excluded from these
statistical comparisons because results were available for only
six datasets.

Table~\ref{tab7} presents the classification accuracy results.
EMG-Hybrid achieved the best mean rank of 1.61, followed by EHHO
(2.89) and BGWOHHO (4.11). It attained the highest classification
accuracy, including tied best results, on 11 of the 14 datasets.
The Friedman test revealed statistically significant differences
among the nine methods ($\chi_F^2=49.422$, $\mathrm{df}=8$,
$p=5.276\times10^{-8}$). Furthermore, all eight pairwise
comparisons between EMG-Hybrid and the baseline methods remained
statistically significant after Holm correction, with adjusted
$p$-values ranging from $9.766\times10^{-4}$ to
$5.493\times10^{-3}$. These findings demonstrate the strong
classification performance of EMG-Hybrid and indicate that its
accuracy advantage is consistent across the evaluated benchmark
datasets.

The feature-subset sizes reported in Table~\ref{tab8} provide
a complementary assessment of feature selection performance.
EHHO achieved the best mean rank (2.14), followed by BGWOHHO
(2.57), while EMG-Hybrid obtained a mean rank of 3.14. The
Friedman test confirmed statistically significant differences
among the methods ($\chi_F^2=57.200$, $\mathrm{df}=8$,
$p=1.646\times10^{-9}$). The Holm-adjusted Wilcoxon tests
identified significant differences between EMG-Hybrid and
HHO ($p_{\mathrm{adj}}=1.709\times10^{-3}$), GA
($p_{\mathrm{adj}}=3.662\times10^{-3}$), and ACO
($p_{\mathrm{adj}}=9.766\times10^{-4}$). However, the
comparisons with MBHO, EHHO, BGWOHHO, GWO, and PSO did not
reach statistical significance after correction. These results
indicate that EMG-Hybrid provides more compact feature subsets
than several conventional baselines, although EHHO and BGWOHHO
achieve better average ranks in terms of feature-subset
compactness.

Regarding optimization fitness (Table~\ref{tab9}), EMG-Hybrid
achieved the best mean rank (2.36), followed by EHHO (2.86)
and BGWOHHO (3.43). It obtained the lowest fitness value on
seven of the 14 datasets. The Friedman test indicated
statistically significant differences among the nine methods
($\chi_F^2=45.333$, $\mathrm{df}=8$,
$p=3.182\times10^{-7}$). After Holm correction, EMG-Hybrid
achieved significantly lower fitness values than MBHO
($p_{\mathrm{adj}}=0.0403$), HHO
($p_{\mathrm{adj}}=0.0093$), GWO
($p_{\mathrm{adj}}=0.0073$), PSO
($p_{\mathrm{adj}}=0.0073$), GA
($p_{\mathrm{adj}}=0.0017$), and ACO
($p_{\mathrm{adj}}=0.00098$). In contrast, the differences
between EMG-Hybrid and EHHO
($p_{\mathrm{adj}}=0.1937$) and between EMG-Hybrid and
BGWOHHO ($p_{\mathrm{adj}}=0.1570$) were not statistically
significant. Overall, EMG-Hybrid demonstrated competitive
optimization performance, with statistically supported
improvements over six of the eight baseline methods.

Forward Selection was additionally evaluated on the six
datasets for which results were available. EMG-Hybrid achieved
higher classification accuracy on all six datasets. In terms
of feature-subset size, Forward Selection produced smaller
subsets on five datasets, whereas EMG-Hybrid selected fewer
features on the breast cancer original dataset (2.61 versus
3.67). For optimization fitness, EMG-Hybrid obtained lower
values on five of the six datasets, while Forward Selection
achieved a lower value on wine (0.0478 versus 0.0569).
The largest absolute fitness difference in favor of EMG-Hybrid
occurred on statlog heart (0.0616 versus 0.1330). These
comparisons illustrate the trade-off between classification
performance and feature-subset compactness. In particular,
EMG-Hybrid consistently achieved higher classification
accuracy and generally better optimization fitness, whereas
Forward Selection tended to produce smaller feature subsets
on the evaluated datasets.

\begin{table*}[!t]
\centering
\caption{Classification accuracy achieved by EMG-Hybrid and baseline feature selection methods across the benchmark datasets. Higher values indicate better classification performance. Forward Selection results are reported only for datasets with available results.}
\label{tab7}
\footnotesize
\setlength{\tabcolsep}{5pt}
\renewcommand{\arraystretch}{1.12}

\resizebox{\textwidth}{!}{%
\begin{tabular}{@{}lcccccccccc@{}}
\toprule
\textbf{Dataset} & \textbf{EMG-Hybrid} & \textbf{MBHO}& \textbf{EHHO} & \textbf{BGWOHHO}
& \textbf{HHO} & \textbf{GWO} & \textbf{PSO}
& \textbf{GA} & \textbf{ACO} & \textbf{Forward} \\
\midrule
breast cancer original & 0.975&0.922 &0.968 & 0.965 & \textbf{0.980} & 0.968
& \textbf{0.980} & 0.963 & 0.963 & 0.969 \\
wine & \textbf{0.993}& 0.982&0.983 & 0.986 & 0.978 & 0.973 & 0.982
& 0.982 & 0.974 & 0.966 \\
statlog heart & \textbf{0.959}& 0.883& 0.931&0.912 & 0.840 & 0.827 & 0.847
& 0.846 & 0.839 & 0.822 \\
zoo & \textbf{0.990} & \textbf{0.99} & 0.983 & 0.972 & \textbf{0.990} & 0.977
& 0.984 & 0.963 & 0.980 & 0.951 \\
vehicle & \textbf{0.814} & 0.734 &0.737& 0.766 & 0.716 & 0.707 & 0.714
& 0.717 & 0.712 & 0.721  \\
waveform & \textbf{0.850}&0.819 &0.837 & 0.817 & 0.801 & \textbf{0.850}
& 0.805 & 0.797 & 0.798 & 0.829 \\
HCV & \textbf{0.408}& 0.322& 0.308 & 0.323 & 0.285 & 0.282
& 0.299 & 0.297 & 0.289 & -- \\
breast cancer diagnostic & \textbf{0.986} & 0.978 & 0.979 &0.975 & 0.971 & 0.969
& 0.975 & 0.970 & 0.969 & --  \\
ionosphere & \textbf{0.962} & 0.926 &0.937 & 0.931 & 0.878 & 0.873
& 0.925 & 0.891 & 0.881 & -- \\
darwin & 0.969& 0.833  &\textbf{0.979} & 0.925 & 0.879 & 0.978
& 0.856 & 0.922 & 0.936 & -- \\
toxicity & \textbf{0.843} &0.678 &0.819 & 0.763 & 0.737 & 0.735
& 0.725 & 0.728 & 0.766 & -- \\
micro-mass & \textbf{0.870} & 0.690 &0.855 & 0.708 & 0.679 & 0.821
& 0.689 & 0.766 & 0.759 & -- \\
leukemia & 0.937 & \textbf{0.969} &0.934 & 0.967 & 0.958 & 0.917
& 0.889 & 0.876 & 0.889 & -- \\
GCM & \textbf{0.709} & 0.684 &0.701 & 0.629 & 0.585 & 0.668
& 0.606 & 0.584 & 0.621 & -- \\

\midrule
\textbf{Mean Rank}
& $\mathbf{1.61}$ & 4.79 & 2.89 & 4.11
& 6.04 & 6.25 & 5.82 & 6.82 & 6.68 & -- \\

\textbf{Friedman Test}
& \multicolumn{9}{c}{$\chi_F^2=49.422$, $\mathrm{df}=8$,
$p=5.276\times10^{-8}$}
& -- \\

\midrule
\textbf{Raw Wilcoxon $p$}
& --
& $2.441\times10^{-3}$
& $1.831\times10^{-3}$
& $1.221\times10^{-3}$
& $2.441\times10^{-3}$
& $7.324\times10^{-4}$
& $2.441\times10^{-4}$
& $1.221\times10^{-4}$
& $1.221\times10^{-4}$
& -- \\

\textbf{Holm-adjusted $p$}
& --
& $5.493\times10^{-3}$
& $5.493\times10^{-3}$
& $4.883\times10^{-3}$
& $5.493\times10^{-3}$
& $3.662\times10^{-3}$
& $1.465\times10^{-3}$
& $9.766\times10^{-4}$
& $9.766\times10^{-4}$
& -- \\

\bottomrule

\end{tabular}%
}
\end{table*}

\begin{table*}[!t]
\centering
\caption{Mean number of selected features obtained by EMG-Hybrid and the baseline feature selection methods across the benchmark datasets. Lower values indicate more compact feature subsets. Forward Selection results are reported only for datasets with available results.}
\label{tab8}
\footnotesize
\setlength{\tabcolsep}{5pt}
\renewcommand{\arraystretch}{1.12}

\resizebox{\textwidth}{!}{%
\begin{tabular}{@{}lcccccccccc@{}}
\toprule
\textbf{Dataset} & \textbf{EMG-Hybrid} & \textbf{MBHO} & \textbf{EHHO} & \textbf{BGWOHHO}
& \textbf{HHO} & \textbf{GWO} & \textbf{PSO}
& \textbf{GA} & \textbf{ACO} & \textbf{Forward} \\
\midrule
breast cancer original
& 2.61& \textbf{2.53}&2.56 & 2.60 & 5.43 & 4.80 & 3.10 & 2.97 & 4.00 & 3.67 \\
wine
& 3.60 &3.53 & \textbf{3.40} & 3.67 & 7.60 & 8.87 & 5.40 & 5.80 & 7.13 & 2.67 \\
statlog heart
& 4.98 & 5.17 &4.63 & 3.86 & 7.90 & 9.50 & 5.83 & 6.40 & 8.33 & \textbf{2.67} \\
zoo
& 6.37&7.36 &8.47 & 7.06 & 9.17 & 6.20 & 10.97 & 8.60 & 11.37 & \textbf{4.67} \\
vehicle
& 7.83 & 7.30& 6.43 & 6.80 & 9.73 & 12.40 & 8.90 & 8.73 & 12.03 & \textbf{5.33} \\
waveform
& 11.47 & 11.59 &9.42 & \textbf{8.67} & 12.87 & 15.43 & 13.03 & 11.37 & 14.23 & 11.00 \\
HCV
& 10.58 & 10.16 &\textbf{3.5} & 10.69 & 14.57 & 19.63 & 7.00 & 10.23 & 17.30 & -- \\
breast cancer diagnostic
& 9.63 & 6.97& 4.5 & \textbf{4.40} & 15.27 & 20.17 & 7.87 & 10.85 & 16.70 & -- \\
ionosphere
& 5.97 &9.64 &\textbf{3.84} & 3.93 & 16.13 & 22.20 & 8.10 & 11.45 & 19.27 & -- \\
darwin
& 46.70& 283 &56.43 & 56.33 & \textbf{46.16} & 92.27 & 203.17 & 182.50 & 263.83 & -- \\
toxicity
& 131.74&601.6 & 118.87 & \textbf{65.33} & 254.23 & 194.17 & 385.53 & 419.17 & 548.16 & -- \\
micro-mass
& \textbf{182.43}& 650.3&370.33 & 357.67 & 402.57 & 283.66 & 394.03 & 511.67 & 629.30 & -- \\
leukemia
& 1271.87 & 3462.63 &\textbf{739.8} & 954.80 & 1488.60 & 753.67 & 2080.47 & 3194.50 & 3085.83 & -- \\
GCM
& \textbf{943.01}& 3205.29 &1007.8 & 1054.33 & 1455.50 & 2745.17 & 4020.53 & 5743.37 & 6694.33 & -- \\

\midrule
\textbf{Mean Rank}
& $\mathbf{3.14}$ & 5.14 & 2.14 & 2.57
& 6.14 & 6.43 & 5.57 & 5.93 & 7.93 & -- \\

\textbf{Friedman Test}
& \multicolumn{9}{c}{$\chi_F^2=57.200$, $\mathrm{df}=8$,
$p=1.646\times10^{-9}$}
& -- \\

\midrule
\textbf{Raw Wilcoxon $p$}
& --
& $5.798\times10^{-2}$
& $5.016\times10^{-1}$
& $6.257\times10^{-1}$
& $2.441\times10^{-4}$
& $1.343\times10^{-2}$
& $1.074\times10^{-2}$
& $6.104\times10^{-4}$
& $1.221\times10^{-4}$
& -- \\

\textbf{Holm-adjusted $p$}
& --
& $1.740\times10^{-1}$
& $1.000$
& $1.000$
& $1.709\times10^{-3}$
& $5.371\times10^{-2}$
& $5.371\times10^{-2}$
& $3.662\times10^{-3}$
& $9.766\times10^{-4}$
& -- \\

\bottomrule

\end{tabular}%
}
\end{table*}

\begin{table*}[htbp]
\centering
\caption{Fitness values obtained by EMG-Hybrid and baseline feature selection methods across the benchmark datasets. Lower values indicate better optimization performance. Forward Selection results are reported only for datasets with available results and are excluded from statistical comparisons.}
\label{tab9}
\footnotesize
\setlength{\tabcolsep}{5pt}
\renewcommand{\arraystretch}{1.12}

\resizebox{\textwidth}{!}{%
\begin{tabular}{@{}lcccccccccc@{}}
\toprule
\textbf{Dataset} & \textbf{EMG-Hybrid}&  \textbf{MBHO}& \textbf{EHHO} & \textbf{BGWOHHO}
& \textbf{HHO} & \textbf{GWO} & \textbf{PSO}
& \textbf{GA} & \textbf{ACO} & \textbf{Forward} \\
\midrule
breast cancer original
& 0.0506 &0.0498  &0.0401& \textbf{0.0382} & 0.0546 & 0.0520 & 0.0409 & 0.0628 & 0.0610 & 0.0527 \\

wine
& 0.0569 &0.0593  &\textbf{0.0318 }& 0.0323 & 0.0622 & 0.0862 & 0.0424 & 0.0496 & 0.0717 & 0.0478 \\

statlog heart
& \textbf{0.0616} & 0.125 &0.0854 & 0.0957 & 0.1307 & 0.2211 & 0.1527 & 0.1231 & 0.1603 & 0.1330 \\

zoo
& 0.0410 &\textbf{0.0394}& 0.0493 & 0.0637 & 0.0590 & 0.0533 & 0.0693 & 0.0588 & 0.0835 & 0.0493 \\

vehicle
& 0.1777&0.284& 0.1820 & \textbf{0.1754} & 0.2857 & 0.2834 & 0.2214 & 0.2898 & 0.2915 & 0.2439 \\

waveform
& \textbf{0.1493}& 0.1738 &0.1698 & 0.1732 & 0.1763 & 0.1511 & 0.2014 & 0.2088 & 0.2284 & 0.1810 \\

HCV
& 0.4435& 0.5458&\textbf{0.4347} & 0.4559 & 0.4839 & 0.6460 & 0.5000 & 0.5722 & 0.4729 & -- \\

breast cancer diagnostic
& 0.0361 & \textbf{0.0260}& 0.0271& 0.0274 & 0.0703 & 0.0684 & 0.0451 & 0.0559 & 0.0657 & -- \\

ionosphere
& \textbf{0.0327}&0.0594 &0.0582 & 0.0652 & 0.1399 & 0.1714 & 0.0736 & 0.1232 & 0.1258 & -- \\

darwin
& \textbf{0.0269} & 0.116 &0.0509 & 0.0659 & 0.1082 & 0.0318 & 0.1490 & 0.1013 & 0.0856 & -- \\

toxicity
& \textbf{0.0774}&0.1818 &0.1529 & 0.1507 & 0.2480 & 0.1546 & 0.1999 & 0.2108 & 0.1807 & -- \\

micro-mass
& \textbf{0.0796}& 0.2178& 0.2290 & 0.2284 & 0.2105 & 0.1345 & 0.2516 & 0.2339 & 0.2202 & -- \\

leukemia
& 0.087 & 0.0480 &0.0621 & \textbf{0.0211} & 0.0447 & 0.0639 & 0.1077 & 0.1462 & 0.0893 & -- \\

GCM
& \textbf{0.1334}&0.336& 0.2954 & 0.3211 & 0.2983 & 0.2330 & 0.3086 & 0.2863 & 0.3398 & -- \\

\midrule
\textbf{Mean Rank}
& $\mathbf{2.36}$ & 4.79 & 2.86 & 3.43
& 6.21 & 5.43 & 6.21 & 6.57 & 7.14 & -- \\

\textbf{Friedman Test}
& \multicolumn{9}{c}{$\chi_F^2=45.333$, $\mathrm{df}=8$,
$p=3.182\times10^{-7}$}
& -- \\

\midrule
\textbf{Raw Wilcoxon $p$}
& --
& $1.343\times10^{-2}$
& $1.937\times10^{-1}$
& $7.849\times10^{-2}$
& $2.319\times10^{-3}$
& $1.221\times10^{-3}$
& $1.221\times10^{-3}$
& $2.441\times10^{-4}$
& $1.221\times10^{-4}$
& -- \\

\textbf{Holm-adjusted $p$}
& --
& $4.028\times10^{-2}$
& $1.937\times10^{-1}$
& $1.570\times10^{-1}$
& $9.277\times10^{-3}$
& $7.324\times10^{-3}$
& $7.324\times10^{-3}$
& $1.709\times10^{-3}$
& $9.766\times10^{-4}$
& -- \\

\bottomrule

\end{tabular}%
}
\end{table*}

\subsection{Empirical Analysis of the Search Dynamics}
\label{subsec:search_dynamics}

\begin{figure}
    \centering
    \includegraphics[width=1.0\textwidth]{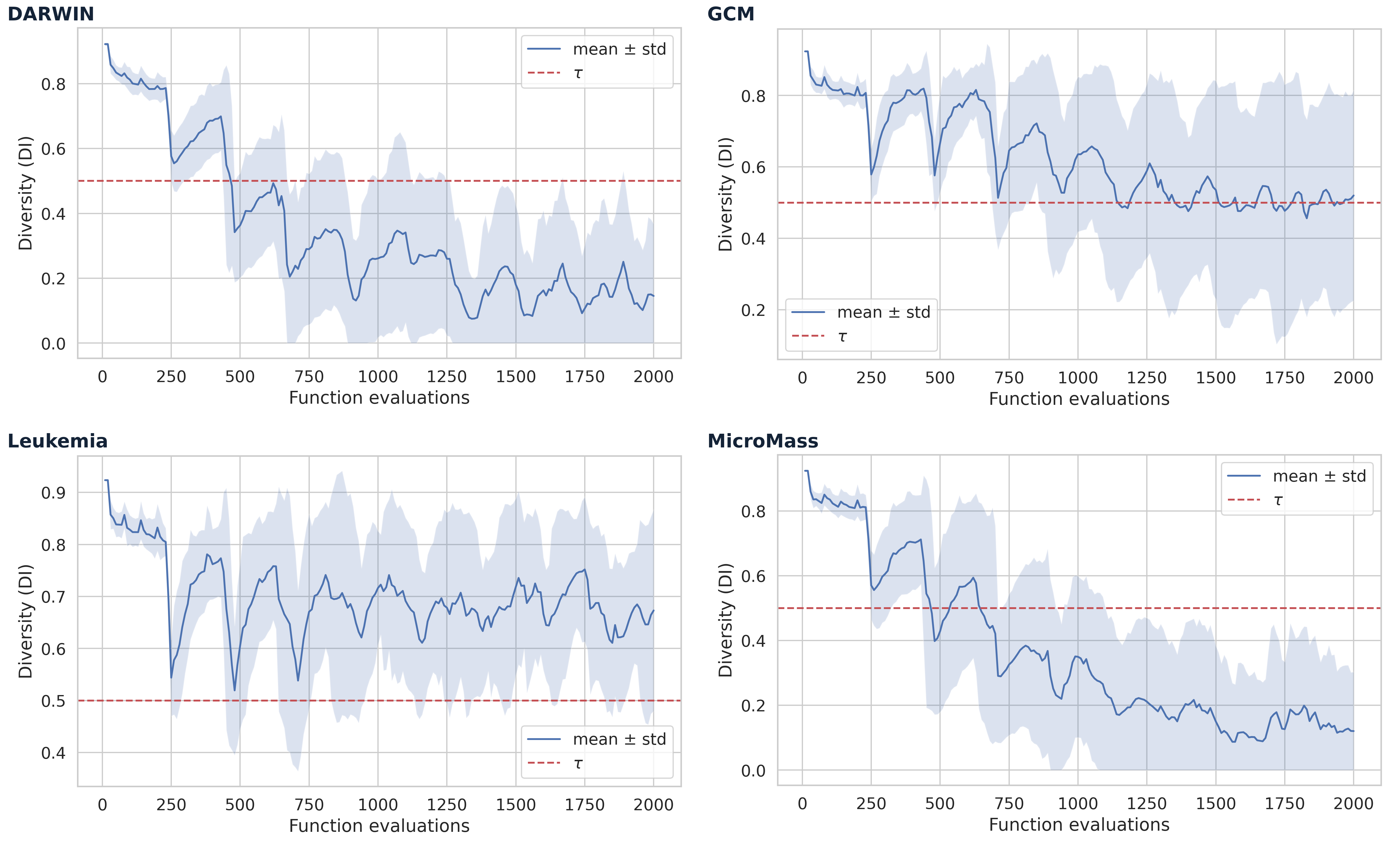}
    \caption{Population diversity (DI) over the function-evaluation budget on the Darwin, GCM, Leukemia, and Micro-mass datasets. The solid curves represent the mean diversity across runs, the shaded regions indicate $\pm$ one standard deviation, and the red dashed lines denote the diversity threshold $\tau$.}
    \label{fig:population_diversity}
\end{figure}

\begin{figure}
    \centering
    \includegraphics[width=1.0\textwidth]{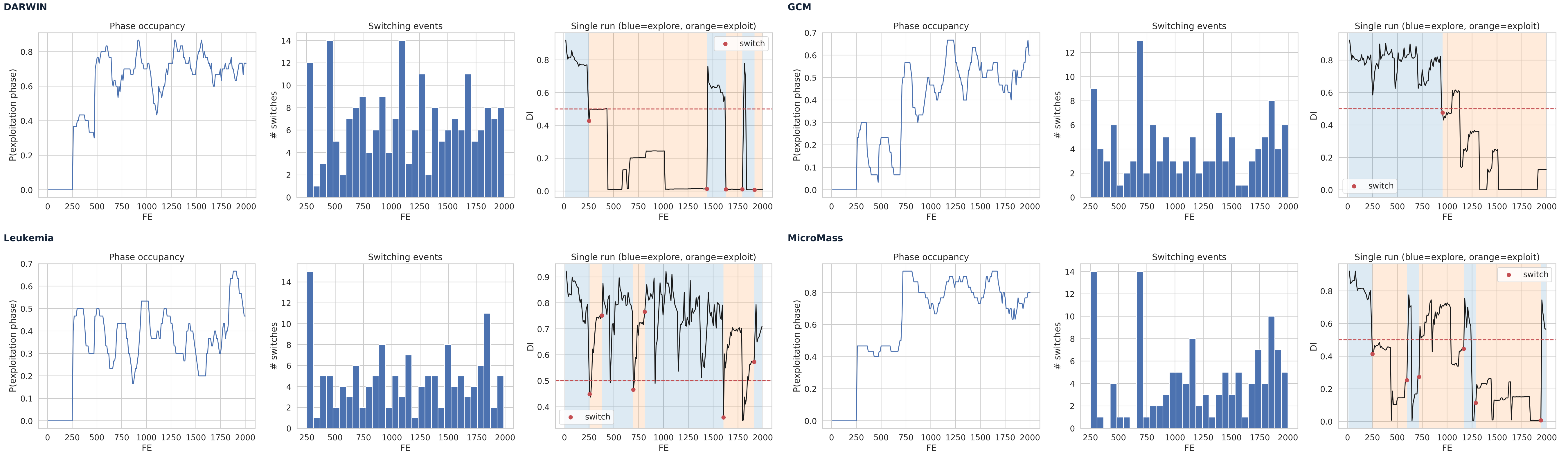}
    \caption{Phase occupancy and switching behavior on the Darwin, GCM, Leukemia, and Micro-mass datasets. For each dataset, the panels show the estimated probability of occupying the exploitation phase, the number of phase-switching events across the function-evaluation budget, and a representative single-run diversity trajectory. In the single-run plots, blue and orange shaded regions indicate the exploration and exploitation phases, respectively; red markers identify switching events, and the dashed red line denotes the diversity threshold $\tau$.}
    \label{fig:switching_events}
\end{figure}

The search dynamics of EMG-Hybrid were examined on four representative datasets: Darwin, GCM, Leukemia, and Micro-mass. Figure~\ref{fig:population_diversity} compares the evolution of population diversity across the function-evaluation budget, including its variability across runs and its relation to the switching threshold $\tau$. Figure~\ref{fig:switching_events} presents the phase-occupancy probabilities, switching frequencies, and representative diversity trajectories, illustrating how the controller regulates transitions between exploration and exploitation. These complementary views provide an empirical characterization of population diversity, improvement activity, and phase-switching behavior across datasets with different characteristics.

\subsection{Ablation}
\label{sec:ablation-main}

An ablation study was conducted on five benchmark datasets to
examine the effects of the main EMG-Hybrid components. The
evaluated variants include single-engine search (HHO only and
GWO only), alternative switching policies (fixed and random),
archive modifications (without DE refinement and single-elite
archive), acceptance strategies (greedy and fixed $\theta$),
and removal of ABHG guidance. All variants followed the same
experimental protocol as the complete framework.

Tables~\ref{tab:ablation} and~\ref{tab:ablation_features}
report the average classification accuracy and number of
selected features, respectively. Full EMG-Hybrid achieved
the highest mean accuracy (0.866), while all ablated variants
obtained lower values, ranging from 0.827 to 0.858. The
largest accuracy reduction occurred with HHO-only search
(0.827), followed by the single-elite archive (0.846).
Alternative switching policies also reduced mean accuracy
to 0.850--0.854, compared with 0.866 for the complete
framework. Removing DE refinement, adaptive threshold
updates, or ABHG guidance resulted in smaller reductions.

The feature-subset results show a different pattern.
Although HHO-only search selected more features on average
(608.3 versus 515.2), GWO-only search and the single-elite
archive produced smaller subsets (465.1 and 499.7,
respectively). These reductions were accompanied by lower
classification accuracy, illustrating the trade-off between
subset compactness and predictive performance. Disabling
DE refinement, adaptive threshold updates, or ABHG guidance
increased the average subset size relative to the complete
framework. Overall, the results indicate that the full
configuration provides the strongest mean classification
accuracy while maintaining a comparatively compact feature
subset. The observed differences are descriptive and do not
establish statistical significance.

\begin{table}[ht]
\centering
\scriptsize
\renewcommand{\arraystretch}{1.15}
\setlength{\tabcolsep}{3pt}
\caption{Ablation study of EMG-Hybrid: average classification accuracy of each variant.}
\label{tab:ablation}

\begin{tabularx}{\textwidth}{
@{}
>{\raggedright\arraybackslash}p{3.2cm}
*{7}{>{\centering\arraybackslash}X}
@{}
}
\toprule
\textbf{Variant} &
\textbf{Darwin} &
\textbf{Toxicity} &
\textbf{Micro-mass} &
\textbf{Leukemia} &
\textbf{GCM} &
\textbf{Mean} &
\textbf{vs. Full} \\
\midrule
Full EMG-Hybrid & 0.969 & 0.843 & 0.870 & 0.937 & 0.709 & 0.866 & -- \\
\midrule
HHO only  & 0.938 & 0.806 & 0.803 & 0.920 & 0.666 & 0.827 & $-4.5\%$ \\
GWO only& 0.972 & 0.805 & 0.853 & 0.917 & 0.695 & 0.848 & $-2.1\%$ \\
Fixed switching  & 0.966 & 0.833 & 0.858 & 0.913 & 0.699 & 0.854 & $-1.4\%$ \\
Random switching & 0.963 & 0.829 & 0.854 & 0.911 & 0.695 & 0.850 & $-1.8\%$ \\
w/o DE on archive & 0.967 & 0.838 & 0.865 & 0.915 & 0.703 & 0.858 & $-0.9\%$ \\
Single-elite archive ($k=1$) & 0.961 & 0.823 & 0.848 & 0.907 & 0.689 & 0.846 & $-2.3\%$ \\
Greedy acceptance ($\theta=0$) & 0.965 & 0.831 & 0.856 & 0.912 & 0.697 & 0.852 & $-1.6\%$ \\
Fixed $\theta$ (no $\mathcal{M}$-Engine) & 0.968 & 0.838 & 0.864 & 0.915 & 0.703 & 0.858 & $-0.9\%$ \\
w/o ABHG ($p_{\mathrm{ABHG}}=0$) & 0.965 & 0.836 & 0.862 & 0.914 & 0.700 & 0.855 & $-1.3\%$ \\
\bottomrule
\end{tabularx}
\end{table}

\begin{table*}
\centering
\caption{Ablation study: average number of selected features across five benchmark datasets.}
\label{tab:ablation_features}
\small
\setlength{\tabcolsep}{5pt}
\renewcommand{\arraystretch}{1.15}
\begin{tabular}{@{}lrrrrrrr@{}}
\toprule
\textbf{Variant} &
\textbf{Darwin} &
\textbf{Toxicity} &
\textbf{Micro-mass} &
\textbf{Leukemia} &
\textbf{GCM} &
\textbf{Mean} &
\textbf{vs. Full} \\
\midrule
Full EMG-Hybrid & 46.7 & 131.7 & 182.4 & 1271.9 & 943.0 & 515.2 & -- \\
\midrule
HHO only                  & 51.4 & 150.2 & 213.4 & 1551.7 & 1075.0 & 608.3 & +18\% \\
GWO only                  & 43.4 & 119.9 & 166.0 & 1119.2 & 877.0 & 465.1 & $-10\%$ \\
Fixed switching           & 48.1 & 135.7 & 187.9 & 1310.0 & 971.3 & 530.6 & +3\% \\
Random switching          & 49.0 & 138.3 & 191.6 & 1335.5 & 990.2 & 540.9 & +5\% \\
w/o DE on archive         & 47.4 & 133.7 & 185.2 & 1291.0 & 957.2 & 522.9 & +1.5\% \\
Single-elite ($k=1$)      & 45.3 & 127.8 & 177.0 & 1233.7 & 914.7 & 499.7 & $-3\%$ \\
Greedy ($\theta=0$)       & 45.8 & 129.1 & 178.8 & 1246.4 & 924.1 & 504.8 & $-2\%$ \\
Fixed $\theta$            & 47.2 & 133.1 & 184.3 & 1284.6 & 952.4 & 520.3 & +1\% \\
w/o ABHG                  & 48.5 & 137.1 & 193.4 & 1348.1 & 980.8 & 541.6 & +5\% \\
\bottomrule
\end{tabular}
\end{table*}

\FloatBarrier

\section{Conclusion and Future Work}
\label{sec6}

This paper introduced EMG-Hybrid, a three-level hierarchical metaheuristic for wrapper-based feature selection. At the control level, the entropy-based $\Lambda$-Engine switches between BHHO exploration and BGWO exploitation, while the $\mathcal{M}$-Engine adjusts the acceptance threshold according to elite fitness. At the search level, ABHG guides binary assimilation toward high-quality solutions. At the memory level, a write-protected elite archive preserves diverse solutions and periodically refines them using discrete differential evolution (DE).

EMG-Hybrid was evaluated on 14 benchmark datasets against eight metaheuristic baselines using leakage-free nested cross-validation. It achieved the best mean ranks in classification accuracy (1.61) and optimization fitness (2.36). Accuracy differences were statistically significant against all eight baselines after Holm correction. For fitness, significant differences were found against six baselines, while comparisons with EHHO and BGWOHHO were not significant. However, EHHO and BGWOHHO achieved better mean ranks for subset size (2.14 and 2.57, respectively) than EMG-Hybrid (3.14). Significant subset-size differences were observed only against HHO, GA, and ACO. Compared with Forward Selection on six datasets, EMG-Hybrid achieved higher accuracy on all six and better fitness on five, although Forward Selection generally selected fewer features. Overall, these results demonstrate strong classification and optimization performance, but not consistently superior subset compactness.

The ablation study examined engine switching, diversity-driven transitions, archive guidance, adaptive threshold adjustment, and ABHG. Removing individual components generally reduced mean classification accuracy, although their effects varied, supporting their contributions to the overall framework.

Several limitations remain. Parameter sensitivity was examined on only five datasets using one-factor-at-a-time analysis, leaving parameter interactions unexplored. Performance was also dataset-dependent, with competing methods outperforming EMG-Hybrid on some benchmarks, including leukemia. Furthermore, the evaluation relied on a $k$-NN wrapper and a small population ($N=10$), limiting conclusions about generalizability to other classifiers and larger populations.

Future work will explore additional classifiers and larger-scale or streaming datasets, develop a multi-objective formulation that separately optimizes classification performance and subset size, investigate adaptive archive management, and theoretically analyze the entropy-based switching criterion.

\subsection*{Conflict of Interest}
The authors declare that they have no conflict of interest.

\subsection*{Data availability}
The datasets analyzed during the current study are publicly available from the UCI Machine Learning Repository (\url{https://archive.ics.uci.edu/}) and the OpenML platform (\url{https://www.openml.org/}). These datasets are open access and can be freely retrieved from the respective websites. 

\subsection*{Funding sources}
This research did not receive any specific grant from funding agencies in the public, commercial, or not-for-profit sectors.

\subsection*{Author Declaration on the Use of AI-Assisted Writing Tools}
The authors declare that during the preparation of this manuscript, they utilized ChatGPT to assist with sentence structure and grammar refinement. Following the use of this tool, the authors thoroughly reviewed and edited the content, and accept full responsibility for the integrity and accuracy of the publication’s final version.

\bibliographystyle{model1-num-names}
\bibliography{cas-refs}

\end{document}